\documentclass[sigconf]{acmart}

\setcopyright{none}              
\renewcommand\footnotetextcopyrightpermission[1]{} 

\usepackage{tabularx}
\newcommand{\Ind}{\mathbb{I}}
\usepackage{multirow}
\AtBeginDocument{%
  }

\begin{document}

\title{SVR: Self-Verifying Refinement via Joint Verdict–Confidence Reinforcement Learning for Adaptive Test-Time Compute}

\author{
  Hongyu Chen\textsuperscript{1},
  Liang Lin\textsuperscript{1,2,3},
  Guangrun Wang\textsuperscript{1,2,3,*}
  \\
  \small{
    \textbf{Email:} chenhy527@mail2.sysu.edu.cn, linliang@live.com, wanggrun@gmail.com
  }
}

\affiliation{
    \institution{
        \textsuperscript{1}Sun Yat-sen University; \textsuperscript{2}Guangdong Key Laboratory of Big Data Analysis and Processing;\textsuperscript{3}X-Era AI Lab
    }
    \city{}
    \country{}
}

\renewcommand{\shortauthors}{Chen et al.}

\begin{abstract}
Scaling test-time computation can improve language-model reasoning, but uniform budgets waste computation on easy inputs, while verifier-guided refinement relies on external feedback. We introduce Self-Verifying Refinement (SVR), an oracle-free multi-turn reinforcement learning framework that learns to use self-verification as a compute-control policy. At each turn, the model produces a solution together with a discrete correctness verdict and a confidence score; it retains the current answer only when the verdict is Correct and confidence exceeds a threshold, and otherwise continues refinement using its own self-verification. Ground-truth correctness is used only to construct training rewards and is never exposed to the policy through refinement prompts or required at inference. SVR is trained with GRPO on fixed-horizon trajectories using rewards that promote solution correctness, calibration-aware self-verification, and stop-ready correct states; adaptive stopping is activated only at inference. On seven mathematical reasoning benchmarks with Qwen3.5-2B, SVR achieves a macro-average accuracy of 0.563 with only 2.99 inference turns on average. In the evaluated complete-system comparison, it exceeds standard GRPO, strong multi-turn baselines, and a fixed-budget oracle-guided score-feedback reference while requiring substantially fewer turns than fixed ten-turn inference. These results demonstrate that learned self-verification can serve as an effective internal control signal for answer retention and adaptive test-time compute allocation.
\end{abstract}





\maketitle
\fancyhead[LE,RO]{}
\begingroup
  \renewcommand\thefootnote{\fnsymbol{footnote}} 
  \footnotetext[2]{Corresponding author: Guangrun Wang}
\endgroup

\section{Introduction}

Scaling test-time computation has become a central strategy for improving language-model reasoning. Models can spend additional inference compute on sampling, search, or iterative refinement rather than committing to a single response \citep{wei2022chain,wang2022selfconsistency,yao2023tree,shinn2023reflexion}. Yet the value of additional computation is highly instance-dependent: an easy problem may already be solved after one attempt, whereas a difficult problem may require several revisions. Additional refinement is also not uniformly beneficial, because a later turn may repair an incorrect solution or overwrite an answer that was already correct. Test-time reasoning is therefore not only a scaling problem, but also a problem of allocating computation and retaining the right intermediate answer.

\begin{figure}[t]
\centering
\includegraphics[width=\columnwidth]{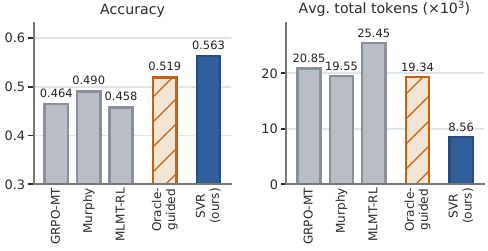}
\vspace{-11pt}
\caption{Accuracy and cumulative inference tokens on All-7 for representative multi-turn refinement methods. Fixed-budget methods execute ten turns, the oracle-guided reference additionally uses external correctness feedback, and SVR performs adaptive stopping. Higher accuracy and lower token consumption indicate better performance.}
\label{fig:accuracy_token_comparison}
\vspace{-11pt}
\end{figure}

Existing approaches address only part of this problem. Fixed-budget methods assign the same number of samples, search steps, or refinement turns to every input despite variation in the marginal value of additional computation \citep{alomrani2025reasoningbudget}. Adaptive allocation methods instead assign different compute budgets across inputs, but typically make an input-level allocation decision before reasoning rather than deciding turn by turn whether to retain or refine the current answer \citep{zhai2026adaptivecompute}. Self-correction methods primarily study how to revise a previous response while leaving the amount of refinement externally specified \citep{shinn2023reflexion,chen2025sets}. Verifier-guided methods can inform selection or revision through reward models, process verifiers, execution results, or correctness checks \citep{cobbe2021training,snell2024scaling,kamoi2025stepverifier}, but these external signals may be costly or unavailable at deployment. Without an internal signal for controlling refinement and answer retention, extending a trajectory may waste tokens and expose correct intermediate answers to harmful revision \citep{sui2025efficientreasoning,zhou2026overthinking}. This raises the central question of our work: can a model learn to retain or refine its current answer using only signals generated by the policy itself, without access to correctness feedback at inference?

A natural candidate is the model's self-verification of its current answer. Once this signal controls computation, however, calibration errors become allocation errors: underconfidence can expose an already correct answer to unnecessary and potentially harmful revision, whereas overconfidence can terminate an incorrect trajectory before additional computation repairs it. Language models are known to be miscalibrated \citep{guo2017calibration,jiang2021calibration}; although they can exhibit meaningful self-evaluation signals \citep{kadavath2022language}, aligning verbalized confidence with answer correctness remains challenging for reasoning models \citep{yoon2025reasoningconfidence,stangel2025rewardingdoubt,liu2026rlcm}. This motivates treating structured self-verification not merely as a post-hoc description of uncertainty, but as a policy-level signal for deciding whether to retain the current answer or allocate another refinement turn.

Based on this premise, we propose \textbf{Self-Verifying Refinement} (SVR), an oracle-free multi-turn reinforcement learning framework that learns to refine answers and generate structured self-verification for inference-time control. Each turn produces a solution, a categorical verdict---\textsc{Correct}, \textsc{Incorrect}, or \textsc{Unsure}---and a confidence estimate of the probability that the current answer is correct. Before reaching the maximum number of turns, SVR stops and returns the current answer only when the verdict is \textsc{Correct} and confidence exceeds a threshold; otherwise, it constructs the next prompt from the original problem, the previous solution, and the policy's own self-verification. Ground-truth correctness is used to construct training rewards but is never exposed in refinement prompts or required by the inference-time controller. Thus, ``oracle-free'' describes the information boundary of the refinement policy rather than the absence of correctness supervision during training.

SVR is trained on fixed-horizon refinement trajectories using Joint Verdict–Confidence Reinforcement Learning, a multi-turn objective based on Group Relative Policy Optimization (GRPO) \citep{shao2024deepseekmath}. Per-turn solve, self-verification, and format scores are averaged into one trajectory return, which supervises only the final-completion tokens; intermediate turns contribute to the return and provide refinement context but are not independent optimization samples. The self-verification terms align confidence with answer correctness through a Brier-style objective \citep{brier1950verification}, penalize overconfident errors, and promote error recognition and correct states eligible for stopping. Because all training trajectories have a fixed horizon, the objective does not directly penalize realized turn or token usage. Compute savings arise at inference, where the verdict--confidence stopping rule can terminate a trajectory before the maximum number of turns.

We evaluate SVR with Qwen3.5-2B on seven mathematical reasoning benchmarks spanning arithmetic search, grade-school word problems, competition mathematics, and multi-step reasoning. SVR reaches an All-7 macro-average accuracy of 0.563 with 2.99 inference turns on average (Figure~\ref{fig:accuracy_token_comparison} and Table~\ref{tab:main_results}). It outperforms the strongest non-oracle multi-turn baseline by 7.5 percentage points and scores 4.4 points above the evaluated fixed-budget oracle-guided reference in a complete-system comparison. Fixed-budget sweeps using the same trained policies show that no single shared stopping turn matches adaptive SVR, supporting the interpretation that its advantage comes from instance-dependent answer retention rather than simply generating more refinement turns. SVR also matches the aggregate accuracy of ten-sample GRPO majority voting while consuming approximately half as many tokens. Together, these results support learned self-verification as an internal signal for both answer retention and adaptive test-time compute allocation.

Our contributions are threefold. First, we introduce SVR, an oracle-free closed-loop refinement framework that uses a policy-generated verdict and confidence estimate to decide whether to retain or revise the current answer. Second, we develop Joint Verdict–Confidence Reinforcement Learning, a fixed-horizon, multi-turn objective that aggregates turn-level solution, self-verification, and formatting signals into a trajectory-level return. This objective trains the policy to produce a verdict–confidence signal for inference-time stopping without directly optimizing the realized inference cost. Third, across seven reasoning benchmarks, SVR achieves a stronger aggregate accuracy--compute trade-off than the evaluated baselines and matches ten-sample GRPO majority voting at approximately half the token cost.

\section{Related Work}

\textbf{Adaptive test-time compute.}\quad Increasing test-time computation can improve language-model reasoning through explicit intermediate reasoning, repeated sampling, and structured search. Chain-of-thought prompting elicits intermediate reasoning steps \citep{wei2022chain}, self-consistency aggregates independently sampled reasoning paths \citep{wang2022selfconsistency}, and tree-structured methods explore and evaluate alternative reasoning paths before selecting an answer \citep{yao2023tree}. Recent work increasingly treats inference compute as a resource-allocation problem: the marginal utility of additional computation varies across inputs, motivating per-prompt allocation under finite budgets \citep{snell2024scaling,alomrani2025reasoningbudget,zhai2026adaptivecompute}. Learning How Hard to Think, for example, predicts reward distributions for input--budget pairs and uses them to allocate best-of-$k$ samples or route inputs between decoders \citep{damani2025learning}. These methods allocate compute primarily from input-level predictions. Yet the value of continued reasoning can change as a solution evolves: additional computation may waste tokens \citep{sui2025efficientreasoning} and even cause models to abandon previously correct answers \citep{zhou2026overthinking}. SVR instead moves adaptive allocation to the evolving solution state, using policy-generated self-verification to decide whether the current answer warrants another refinement turn.

\noindent\textbf{Multi-turn refinement and reinforcement learning.}\quad Multi-turn reasoning allows a model to revise previous responses instead of treating each problem as a single-shot generation task. Reflexion uses verbal feedback and episodic memory to guide subsequent attempts \citep{shinn2023reflexion}, while SCoRe trains language models to self-correct through multi-turn online reinforcement learning \citep{kumar2024score}. SETS combines sampling, self-verification, and self-correction at test time without additional model training \citep{chen2025sets}. MURPHY extends GRPO to feedback-conditioned multi-turn code generation with retrospective credit assignment \citep{ekbote2025murphy}, whereas iGRPO selects a high-reward model-generated draft and trains a draft-conditioned refinement in a second optimization stage \citep{hatamizadeh2026igrpo}. In parallel, reinforcement learning with verifiable rewards has become a major approach to mathematical reasoning. DeepSeekMath introduced Group Relative Policy Optimization (GRPO) \citep{shao2024deepseekmath}, and systems such as DeepSeek-R1, Open-Reasoner-Zero, and VAPO show that outcome-based reinforcement learning can improve long-chain reasoning and benchmark performance \citep{deepseekai2025deepseekr1,hu2025openreasonerzero,yue2025vapo}. Data-efficient RLVR further shows that substantial reasoning gains can emerge from very limited training examples \citep{wang2025oneexample}. Across these lines, refinement and turn-level control remain separate: multi-turn structures are preset or externally organized, while RLVR optimizes single-response trajectories. SVR integrates self-verification into multi-turn RL, so the same policy learns both to refine answers and to produce the signal used for inference-time control.

\begin{figure*}[t]
\centering
\includegraphics[width=\textwidth]{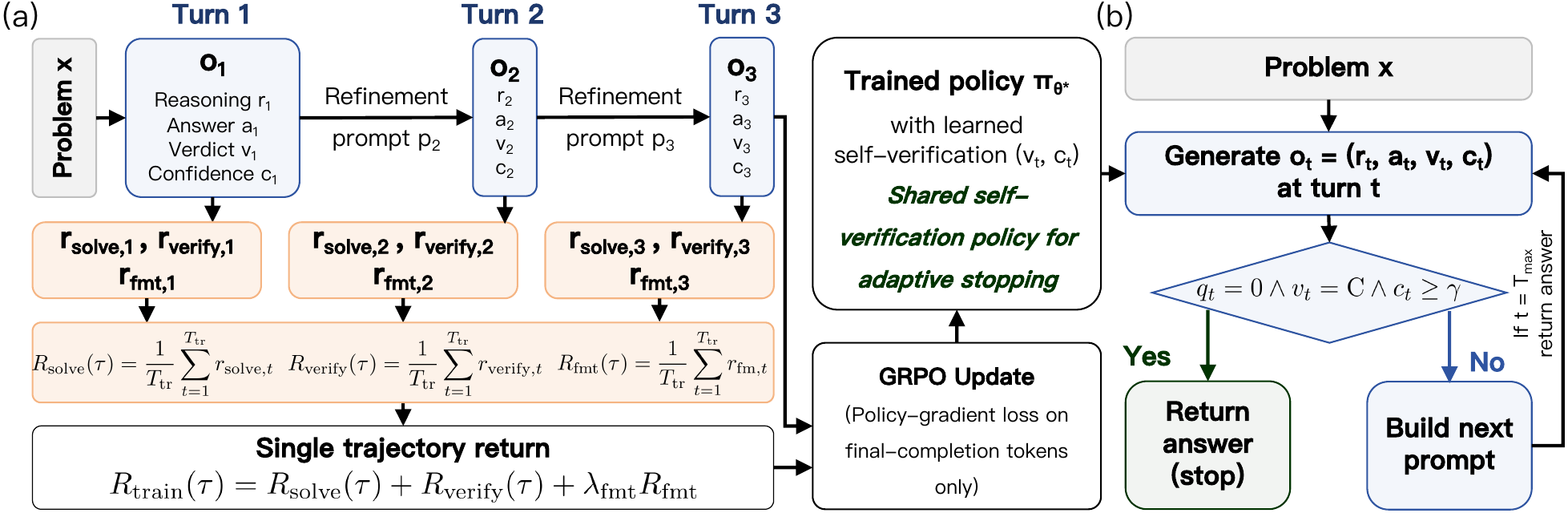}
\vspace{-14pt}
\caption{Overview of fixed-horizon training and adaptive inference in SVR. \textbf{(a)} Per-turn solve, self-verification, and format scores are averaged into one trajectory return, while the policy-gradient loss is applied only to the final-completion tokens; ground-truth correctness is used only for reward construction. \textbf{(b)} Inference stops at the first non-truncated output satisfying $q_t=0$, $v_t=\mathrm{C}$ and $c_t\geq\gamma$; otherwise, SVR continues refine until $T_{\max}$ and returns the final output.}
\label{fig:svr_method}
\vspace{-6pt}
\end{figure*}

\noindent\textbf{Verification and confidence-based control.}\quad Verification provides an important mechanism for improving reasoning quality. Outcome verifiers select final answers from sampled candidates \citep{cobbe2021training}, while process reward models evaluate intermediate reasoning steps \citep{lightman2023process}. Formal tools can also synthesize step-level verification labels for formally checkable tasks, reducing the need for human annotation \citep{kamoi2025stepverifier}. Although effective, these verifier-guided methods generally require an auxiliary model, an execution environment, or a task-specific correctness signal at inference. A related line of work studies confidence calibration. Modern neural networks are often miscalibrated \citep{guo2017calibration}, and confidence estimates for language-model question answering can likewise deviate from empirical correctness \citep{jiang2021calibration}. At the same time, language models can expose informative self-evaluation signals under suitable elicitation formats \citep{kadavath2022language}, and reasoning models often express confidence more accurately than their non-reasoning counterparts \citep{yoon2025reasoningconfidence}. Recent methods directly optimize confidence expression through reinforcement learning or calibration-aware process rewards \citep{stangel2025rewardingdoubt,liu2026rlcm}. Closest to our setting, C3RL jointly optimizes answer correctness and verbalized-confidence calibration, while CAS uses the resulting confidence to allocate additional independent samples toward low-confidence inputs \citep{yang2026scalingconfidence}. CoRefine instead trains a lightweight controller over confidence derived from a frozen model's token-level traces to select among halting and refinement actions \citep{jin2026corefine}. Prior designs either externalize verification and control or use confidence primarily to allocate independent samples. SVR instead embeds a verdict--confidence interface in the reasoning policy itself and stops only when the verdict is \textsc{Correct} and the estimated probability that the current answer is correct exceeds the threshold, without a separate verifier or controller at inference.

\section{Methodology}

\subsection{Adaptive Refinement Formulation}
\label{sec:formulation}

We formulate multi-turn refinement as a sequential test-time compute allocation problem. Given an input \(x\), a policy \(\pi_\theta\) generates a sequence of candidate solutions using at most \(T_{\max}\) inference turns. After each generation, a deterministic controller either returns the current answer or allocates another refinement turn using only the policy-generated self-verification state and generation status. We use \(T_{\mathrm{tr}}\) to denote the fixed rollout horizon during training and \(T_{\max}\) to denote the maximum deployment budget. Figure~\ref{fig:svr_method} contrasts fixed-horizon optimization with adaptive inference.

Let \(p_t\) denote the policy-visible prompt at turn \(t\). The initial prompt is constructed as
\begin{equation}
p_1=\mathcal{P}_0(x),
\label{eq:initial-prompt}
\end{equation}
where \(\mathcal{P}_0\) combines the original problem with the fixed system instruction, task-specific answer convention, and structured self-check requirements. At each turn, the policy generates a completion whose structured representation is
\begin{equation}
o_t=(r_t,a_t,v_t,c_t)\sim\pi_\theta(\cdot\mid p_t),
\label{eq:turn-output}
\end{equation}
where \(r_t\) is the reasoning trace, \(a_t\) is the task answer, and \((v_t,c_t)\) is the parsed self-verification state. The normalized verdict \(v_t\in\mathcal{V}=\{\mathrm{C},\mathrm{I},\mathrm{U}\}\) denotes \textsc{Correct}, \textsc{Incorrect}, or \textsc{Unsure}. The canonical self-check requests a \textsc{Correct} or \textsc{Incorrect} judgment, while explicit uncertainty or an invalid verdict is mapped to \(\mathrm{U}\). The confidence \(c_t\in[0,1]\) is the policy-reported confidence that \(a_t\) is correct. It is not assumed to be calibrated a priori; the objective introduced in Section~\ref{sec:learning-controller} trains it to better reflect empirical answer correctness.

For reward construction and evaluation, we define
\begin{equation}
y_t=\mathbb{I}[a_t\text{ is correct}],
\qquad
q_t=\mathbb{I}[o_t\text{ is truncated}],
\label{eq:turn-state}
\end{equation}
where an unparseable answer is treated as incorrect. The task-specific evaluator supplies \(y_t\in\{0,1\}\) only during training and evaluation, whereas \(q_t\in\{0,1\}\) is ordinary generation metadata available at inference. The inference-time controller observes \(o_t\) and \(q_t\), but never accesses \(y_t\), a reference answer, an evaluator score, or any other external correctness signal.

For a non-truncated output, SVR returns the current answer only when the verdict is \textsc{Correct} and the reported confidence is at least a deployment threshold \(\gamma\); otherwise, it allocates another refinement turn whenever budget remains. The verdict gate prevents confidence alone from triggering termination without an explicit positive assessment, while the confidence threshold controls how conservatively such assessments are accepted. The realized stopping time is
\begin{equation}
\hat{t}_{\gamma}
=
\min\left(
\left\{
t\in\{1,\ldots,T_{\max}\}
:
q_t=0,\;
v_t=\mathrm{C},\;
c_t\geq\gamma
\right\}
\cup
\left\{T_{\max}\right\}
\right).
\label{eq:stopping-time}
\end{equation}
The inclusion of \(\{T_{\max}\}\) guarantees termination when no earlier output satisfies the stopping gate. SVR returns \(a_{\hat{t}_\gamma}\), producing an adaptive trajectory \(\tau_\gamma=(o_1,\ldots,o_{\hat{t}_\gamma})\) whose length depends on the input and is bounded by \(T_{\max}\).

Let \(\mathbb{E}_{\gamma}\) denote expectation over \(x\sim\mathcal{D}\) and the adaptive trajectory \(\tau_\gamma\sim\pi_\theta(\cdot\mid x)\). We characterize deployment behavior by
\begin{equation}
\mathcal{A}(\theta,\gamma)
=
\mathbb{E}_{\gamma}
\left[y_{\hat{t}_\gamma}\right],
\mathcal{C}_{\kappa}(\theta,\gamma)
=
\mathbb{E}_{\gamma}
\left[
\sum_{t=1}^{\hat{t}_\gamma}
\kappa(p_t,o_t)
\right].
\label{eq:accuracy-compute}
\end{equation}
Setting \(\kappa_{\mathrm{turn}}(p_t,o_t)=1\) yields the expected number of executed turns, while \(\kappa_{\mathrm{tok}}(p_t,o_t)=|p_t|_{\mathrm{tok}}+|o_t|_{\mathrm{tok}}\) yields the expected cumulative number of tokenized model-input and completion tokens. Both functionals are estimated empirically by averaging realized stopping outcomes and cumulative costs over evaluation examples.

\subsection{Oracle-Free SVR Policy}
\label{sec:oracle-free-policy}

SVR realizes closed-loop refinement with a single policy that jointly produces a candidate solution and its self-assessment, without querying an auxiliary verifier. At each turn, the controller extracts \(v_t\) and \(c_t\) from the generated self-check. Outputs that do not yield a valid positive assessment are treated conservatively and cannot activate the stopping rule. The serialization, normalization, and fallback rules are provided in Appendix~\ref{app:self_check_parsing}.

Let \(d_t\) denote the length-bounded textual draft retained from the reasoning and answer generated at turn \(t\). Whenever refinement continues, SVR constructs the next policy-visible prompt as
\begin{equation}
p_{t+1}
=
\mathcal{P}_{\mathrm{ref}}
\left(
x,\,
d_t,\,
\mathcal{I}(v_t,c_t,q_t)
\right),
\label{eq:refinement-prompt}
\end{equation}
where \(\mathcal{P}_{\mathrm{ref}}\) combines the fixed system instruction with a newly constructed user message containing the original problem, the retained draft, and a state-dependent refinement instruction. During both training and inference, the verdict--confidence state is therefore used to construct the next prompt whenever continuation occurs.

Generation truncation takes precedence over the parsed self-assessment: a truncated output cannot activate the stopping gate, and continuation requests a fresh complete response whenever budget remains. For a complete output that does not terminate the trajectory, \(v_t=\mathrm{C}\) prompts independent re-examination of the most error-prone reasoning step, whereas \(v_t\in\{\mathrm{I},\mathrm{U}\}\) prompts error localization and correction. The confidence \(c_t\) is exposed as part of the refinement context and participates in the stopping rule of Eq.~\eqref{eq:stopping-time}, but it does not independently select the principal refinement mode, which is determined by \(q_t\) and \(v_t\).

SVR uses a first-order context whose history component is bounded independently of the number of refinement turns. Each prompt depends only on \(x\), the retained draft \(d_t\), and the current parsed state \((v_t,c_t,q_t)\), rather than the complete interaction history \((o_1,\ldots,o_t)\). More importantly, Eq.~\eqref{eq:refinement-prompt} exposes only policy-generated information. Ground-truth labels, reference answers, evaluator scores, execution results, and reward values may be used for training or evaluation, but are never included in refinement prompts or accessed by the inference-time controller. Training and inference therefore share the same oracle-free, policy-visible information interface.

\subsection{Learning the Self-Verification Controller}
\label{sec:learning-controller}

SVR must learn two coupled capabilities: improving the current solution through refinement and determining whether that solution is reliable enough to return. Terminal correctness alone is insufficient for this purpose, because trajectories with the same final outcome may differ substantially in their intermediate behavior: one may correct an earlier error, whereas another may overwrite a valid solution and recover only later. Adaptive inference also depends directly on the reliability of intermediate self-assessments. We therefore train SVR on fixed-horizon trajectories \(\tau=(o_1,\ldots,o_{T_{\mathrm{tr}}})\) and evaluate solution quality, self-verification, and output validity at every turn. The evaluator supplies \(y_t\) only for reward construction.

The trajectory-level return averages three complementary per-turn signals:
\begin{equation}
R_{\mathrm{SVR}}(\tau)
=
\frac{1}{T_{\mathrm{tr}}}
\sum_{t=1}^{T_{\mathrm{tr}}}
\left(
r_{\mathrm{solve},t}
+
r_{\mathrm{verify},t}
+
\lambda_{\mathrm{fmt}}r_{\mathrm{fmt},t}
\right).
\label{eq:svr-reward}
\end{equation}
This allows intermediate states to influence learning rather than serving merely as context for the terminal prediction. It distinguishes productive correction from destructive revision and supervises the verdict--confidence state at positions that adaptive inference may not reach. Averaging prevents the return scale from growing mechanically with the training horizon. The format term encourages both a valid task-answer representation and a parseable self-check. Detailed reward definitions and coefficients are provided in Appendix~\ref{app:reward-details}..

The solution component combines current answer quality with cross-turn progress:
\begin{equation}
r_{\mathrm{solve},t}
=
r_{\mathrm{abs},t}
+
r_{\mathrm{prog},t}
-
\lambda_{\mathrm{trunc}}q_t,
\label{eq:solve-reward}
\end{equation}
where \(r_{\mathrm{abs},t}\) rewards correct solutions with local completion-length shaping, without rewarding short incorrect answers, and \(r_{\mathrm{prog},t}\) evaluates the correctness transition between adjacent turns. We set \(r_{\mathrm{prog},1}=0\), so the initial turn is evaluated only by its absolute solution quality. For subsequent turns, the progress term rewards wrong-to-correct refinement and preservation of an already correct answer, while penalizing regression and repeated failure. This transition-aware design encourages the policy not only to reach a correct solution, but also to make useful revisions and avoid damaging a valid one. A truncated completion is still evaluated according to its parsed answer but receives an additional penalty through \(q_t\).

The central self-verification component trains the verdict--confidence pair to function as an actionable stopping signal:
\begin{equation}
\begin{aligned}
r_{\mathrm{verify},t}
&=
(1-q_t)
\Bigl(
\lambda_{\mathrm{cal}}r_{\mathrm{cal},t}
-\lambda_{\mathrm{over}}r_{\mathrm{over},t}
+\lambda_{\mathrm{detect}}r_{\mathrm{detect},t}
+\lambda_{\mathrm{ready}}r_{\mathrm{ready},t}
\Bigr)\\
&=
(1-q_t)
\Bigl(
\lambda_{\mathrm{cal}}\!\left[1-(c_t-y_t)^2\right]
-\lambda_{\mathrm{over}}c_t\mathbb{I}[v_t=\mathrm{C}\land y_t=0]\\
&\qquad
+\lambda_{\mathrm{detect}}\mathbb{I}[v_t=\mathrm{I}\land y_t=0]
+\lambda_{\mathrm{ready}}c_t\mathbb{I}[v_t=\mathrm{C}\land y_t=1]
\Bigr).
\end{aligned}
\label{eq:self-verification-reward}
\end{equation}
The multiplicative mask excludes truncated generations from self-verification supervision because their verdict and confidence may be incomplete. For complete outputs, the calibration component aligns confidence with empirical correctness through a Brier-style objective. Calibration alone, however, does not capture the asymmetric consequences of controller errors. Confidently accepting an incorrect answer may terminate refinement and directly reduce accuracy, whereas rejecting a correct answer primarily incurs additional computation. The overconfidence component therefore penalizes high-confidence false-positive assessments, the detection component rewards explicit recognition of incorrect states, and the stop-readiness component encourages correct answers to carry sufficiently strong positive assessments for confidence-gated termination. Together, these components train self-verification as a control signal rather than merely a descriptive confidence report.

The deployment threshold \(\gamma\) does not appear in the training objective. SVR instead learns a general verdict--confidence signal from the intermediate states encountered under fixed-horizon refinement, after which \(\gamma\) selects the desired inference-time operating point between conservative continuation and aggressive early stopping. Likewise, the trajectory reward does not directly optimize the realized stopping turn or cumulative prompt--completion cost. Local completion-length shaping discourages unnecessarily verbose individual solutions, while adaptive computation emerges at deployment from applying the learned self-verification signal through Eq.~\eqref{eq:stopping-time}. This separation allows a single trained policy to support different accuracy--compute trade-offs without retraining the policy for each deployment threshold.

\subsection{Fixed-Horizon Optimization and Adaptive Inference}
\label{sec:optimization-inference}

SVR is optimized using Joint Verdict–Confidence Reinforcement Learning over fixed-horizon refinement trajectories. For each input \(x\), the policy samples a group of \(G\) trajectories \(\{\tau_i\}_{i=1}^{G}\) and executes every trajectory for \(T_{\mathrm{tr}}\) turns, irrespective of intermediate verdicts or confidence. Forced continuation decouples training-state coverage from an initially unreliable stopping controller. Otherwise, erroneous positive assessments could terminate trajectories before the policy observes the continuation states needed to improve both refinement and self-verification. Fixed-horizon collection therefore provides uncensored supervision over a common refinement depth, while \(T_{\mathrm{tr}}\) serves as a training horizon rather than a deployment-time stopping budget.

Each trajectory is assigned the return defined in Eq.~\eqref{eq:svr-reward}, and the resulting returns are normalized within the group:
\begin{equation}
R_i=R_{\mathrm{SVR}}(\tau_i),
\qquad
\widehat{A}_i
=
\frac{R_i-\mu_R}{\sigma_R+\epsilon},
\label{eq:group-advantage}
\end{equation}
where \(\mu_R\) and \(\sigma_R\) are the mean and standard deviation of the trajectory returns within the group. The standard clipped GRPO objective is then applied using the group-relative advantage \(\widehat{A}_i\). Joint Verdict–Confidence Reinforcement Learning therefore compares trajectories not only in terms of solution quality and refinement progress, but also in terms of whether their policy-generated verdict–confidence states reliably characterize intermediate correctness.

Although \(R_i\) aggregates solve, self-verification, and format evidence from all \(T_{\mathrm{tr}}\) turns, the policy-gradient loss is applied to all generated tokens of only the final completion \(o_{i,T_{\mathrm{tr}}}\). Earlier outputs influence optimization by determining subsequent refinement contexts and by contributing to the trajectory return, but they are not unpacked into separate turn-level optimization samples. Consequently, each trajectory yields one group-relative optimization sample whose final completion is updated using a return that summarizes the full refinement trajectory.

At inference, SVR retains the learned policy and oracle-free information interface but replaces forced continuation with the adaptive scheduler defined in Eq.~\eqref{eq:stopping-time}. The current answer is returned when the verdict--confidence gate is satisfied; otherwise, the next prompt is constructed through Eq.~\eqref{eq:refinement-prompt} while budget remains. The training horizon \(T_{\mathrm{tr}}\) and deployment budget \(T_{\max}\) are independent design parameters and need not coincide. Training and inference thus differ in scheduling rather than information access: fixed-horizon optimization provides uncensored trajectory supervision, whereas adaptive inference lets the learned verdict--confidence state allocate computation on an instance-specific basis.

\section{Experiments}

Our experiments address three questions. First, does SVR improve the accuracy--compute frontier over single-turn, fixed-budget multi-turn, and oracle-guided references? Second, does its learned stopping policy allocate inference computation more effectively than uniform fixed-turn refinement or independent test-time sampling? Third, which reward and self-check components are responsible for this behavior, and are the resulting gains stable across training seeds?

\subsection{Experimental Setup}
\label{sec:experimental_setup}

\noindent\textbf{Datasets and evaluation.}\quad We use the Qwen3.5-2B as the backbone and train separate domain-specific policies on Countdown, GSM8K, and MATH. Countdown uses $50{,}000$ examples sampled from \path{Jiayi-Pan/Countdown-Tasks-3to4}, while GSM8K and MATH use their complete training splits of $7{,}473$ and $7{,}500$ examples, respectively. The Countdown-trained and GSM8K-trained policies are evaluated on their corresponding held-out sets, whereas the MATH-trained policy is evaluated without further fine-tuning on MATH500~\citep{lightman2023process}, AIME26~\citep{dekoninck2026matharena}, AMC23, OlympiadBench~\citep{he2024olympiadbench}, and MinervaMath~\citep{lewkowycz2022solving}. We report All-7 as the unweighted macro-average over all seven benchmarks and Math-5 as the corresponding average over the five benchmarks evaluated with the MATH-trained policy. All trainable methods use full-parameter reinforcement learning in the ms-swift framework~\citep{zhao2024swiftascalablelightweightinfrastructure}, with colocated vLLM rollout generation on four NVIDIA A800 GPUs.

\begin{table*}[t]
\centering
\small
\setlength{\tabcolsep}{4.0pt}
\caption{Final-answer accuracy across seven reasoning benchmarks. All-7 is the macro-average over all benchmarks, and Math-5 averages the five benchmarks evaluated with the MATH-trained policy. Bold and underlined values denote the best and second-best results, respectively.}
\label{tab:main_results}
\vspace{-11pt}
\resizebox{\textwidth}{!}{
\begin{tabular}{lccccccccc}
\toprule
\textbf{Method}
& \textbf{Countdown}
& \textbf{GSM8K}
& \textbf{MATH500}
& \textbf{AIME26}
& \textbf{AMC23}
& \textbf{Olymp.}
& \textbf{Minerva}
& \textbf{All-7}
& \textbf{Math-5} \\
\midrule

Qwen3.5-2B
& 0.529
& 0.775
& 0.564
& 0.067
& 0.325
& 0.316
& 0.364
& 0.420
& 0.327 \\

GRPO
& 0.585
& 0.748
& 0.572
& 0.100
& 0.275
& 0.307
& 0.401
& 0.427
& 0.331 \\

GSPO
& 0.219
& 0.757
& 0.568
& 0.033
& 0.375
& 0.335
& 0.408
& 0.385
& 0.344 \\

\midrule

GRPO-MT
& 0.620
& 0.810
& 0.598
& 0.133
& 0.350
& 0.349
& 0.390
& 0.464
& 0.364 \\

iGRPO
& 0.526
& 0.769
& 0.586
& 0.067
& 0.250
& 0.309
& 0.401
& 0.415
& 0.322 \\

Murphy
& 0.675
& 0.809
& 0.630
& 0.033
& 0.475
& 0.375
& 0.419
& 0.488
& 0.387 \\

MLMT-RL
& 0.555
& 0.776
& 0.590
& \underline{0.167}
& 0.425
& 0.310
& 0.382
& 0.458
& 0.375 \\

ScRPO
& 0.541
& 0.810
& 0.590
& 0.067
& 0.250
& 0.350
& \underline{0.434}
& 0.434
& 0.338 \\

Oracle-guided (fixed)
& \underline{0.745}
& \textbf{0.829}
& \underline{0.636}
& 0.100
& \underline{0.525}
& \underline{0.377}
& 0.423
& \underline{0.519}
& \underline{0.412} \\

\midrule

\textbf{SVR (ours)}
& \textbf{0.839}
& \underline{0.813}
& \textbf{0.676}
& \textbf{0.200}
& \textbf{0.550}
& \textbf{0.417}
& \textbf{0.449}
& \textbf{0.563}
& \textbf{0.458} \\

\bottomrule
\end{tabular}
}
\vspace{-11pt}
\end{table*}

\noindent\textbf{Baselines and generation-count control.}\quad We compare SVR with the unmodified backbone; single-turn GRPO~\citep{shao2024deepseekmath} and GSPO~\citep{zheng2025gspo}; the fixed-horizon multi-turn baseline GRPO-MT; and iGRPO~\citep{hatamizadeh2026igrpo}, Murphy~\citep{ekbote2025murphy}, MLMT-RL~\citep{singh2026mlmtrl}, and ScRPO~\citep{li2025scrpo}. All trainable methods are optimized for one epoch using the same domain-specific data and task evaluator whenever applicable. To control the dominant rollout-generation count, GRPO and GSPO sample $24$ single-turn completions per input; GRPO-MT, Murphy, MLMT-RL, ScRPO, SVR, and the oracle-guided reference sample eight three-turn trajectories, also yielding $24$ turn-level generations; and iGRPO uses $12$ exploratory drafts followed by $12$ conditioned refinements. This alignment controls generated completions, not the number of policy-gradient samples: SVR returns one optimized final completion per trajectory, so its eight trajectories yield eight optimized samples, while the first two turns provide refinement context and contribute per-turn quantities to the averaged trajectory return. Token-level and wall-clock costs may also differ because multi-turn prompts are constructed sequentially. SVR follows the fixed-horizon protocol in Section~\ref{sec:optimization-inference}, with $G=8$, $T_{\mathrm{tr}}=3$, and adaptive stopping disabled during training. The oracle-guided reference uses the same training horizon, Markov context reset, draft limits, and task-specific answer interface as SVR, but inserts an evaluator-derived score of the previous response into each subsequent prompt instead of generating a verdict--confidence pair. Because it provides no policy-generated stopping signal, it is evaluated as a fixed-budget privileged-feedback reference.

\noindent\textbf{Inference protocols and metrics.}\quad Unless otherwise specified, evaluation uses greedy decoding with temperature zero. Single-turn methods return their first response, fixed-budget multi-turn methods run for ten turns, and SVR performs adaptive inference with $T_{\max}=10$ and a single global threshold $\gamma=0.85$. The same threshold is used throughout the main comparison, adaptive-inference analysis, majority-voting comparison, and ablation study; benchmark-specific optima from the diagnostic threshold sweep are not substituted into the reported main results. The majority-voting experiment is the only exception to greedy decoding: it independently samples $k\in\{3,5,10\}$ single-turn GRPO solutions using temperature $0.6$, top-$p=0.95$, and top-$k=20$, and returns the most frequent normalized answer. We report final-answer accuracy and the mean number of generated turns as the primary effectiveness and inference-cost metrics. For adaptive inference, we additionally report Early Stop Rate (ESR), the fraction of examples terminated before $T_{\max}$ by the confidence gate, and Premature Stop Error (PSE), the fraction of all examples that terminate early with an incorrect returned answer. Complete optimization, generation, reward, dataset, and metric specifications are provided in the appendix.

\subsection{Main Results and Compute Efficiency}
\label{sec:main_results}

Across the aggregate columns of Table~\ref{tab:main_results}, SVR attains the strongest performance, reaching an All-7 accuracy of $0.563$ and a Math-5 accuracy of $0.458$. Relative to the unmodified backbone, these results represent improvements of $14.3$ and $13.1$ percentage points, respectively. SVR also outperforms Murphy, the strongest non-oracle multi-turn baseline, by $7.5$ points on All-7 and $7.1$ points on Math-5, and scores $4.4$ and $4.6$ points above the evaluated fixed-budget oracle-guided reference in this complete-system comparison.

These accuracy gains are obtained with substantially lower inference cost. SVR uses only $2.99$ turns on All-7 and $3.42$ turns on Math-5 on average, compared with the ten-turn budget used by fixed-budget refinement methods. The token-level comparison in Figure~\ref{fig:accuracy_token_comparison} reinforces this result: SVR consumes $8.56$ thousand total tokens per example on All-7, less than half the token consumption of the fixed-budget multi-turn methods included there. The oracle-guided reference receives evaluator-derived feedback but lacks a policy-generated stopping signal; therefore, this result compares the complete refinement systems rather than isolating the feedback source under an otherwise identical controller.

The improvement is broadly distributed across benchmarks. SVR ranks first on six of the seven datasets and on all five benchmarks evaluated with the MATH-trained policy. GSM8K is the only exception, where the oracle-guided reference is higher by $0.7$ percentage points. Because AIME26 and AMC23 contain only $30$ and $40$ examples, respectively, we emphasize the aggregate results and trends on the larger benchmarks.

The aggregate improvements are also stable across training seeds. Across three independently trained checkpoints, SVR obtains an All-7 accuracy of $0.556\pm0.007$ with $3.05\pm0.06$ turns and $8.68\pm0.11$ thousand tokens per example. On Math-5, the corresponding results are $0.449\pm0.008$ accuracy, $3.49\pm0.07$ turns, and $11.21\pm0.16$ thousand tokens. All three checkpoints score above the fixed-budget oracle-guided reference in the reported complete-system comparison; its All-7 and Math-5 accuracies are $0.519$ and $0.412$. Complete seed-level and per-dataset results are provided in Appendix~\ref{app:seed_robustness}. These results support the stability of SVR's accuracy--efficiency behavior, although three runs are insufficient for a formal statistical-significance claim.

\subsection{Analysis of Adaptive Compute Allocation}
\label{sec:adaptive_inference_analysis}

The central question behind adaptive refinement is not whether additional turns can occasionally improve an answer, but whether a single shared budget can preserve the most useful answer for every input. We study this question by comparing adaptive SVR with fixed-budget inference using the same trained policies and greedy decoding. Fixed-budget inference returns the answer produced at a common turn $K\in\{1,\ldots,T_{\max}\}$ for every example, whereas adaptive SVR returns the first answer satisfying the stopping rule in Eq.~\eqref{eq:stopping-time}. Thus, the comparison changes the answer-retention rule rather than the underlying policy or decoding procedure. Complete ten-turn trajectories are used to evaluate the corresponding fixed-budget prefixes, without majority voting, best-of-turn selection, or retrospective oracle selection.

\begin{figure}[h]
\centering
\vspace{-3pt}
\includegraphics[width=\columnwidth]{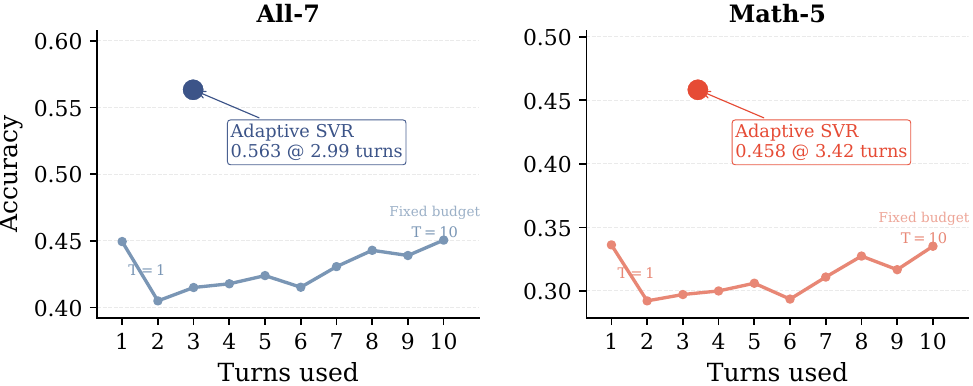}
\vspace{-14pt}
\caption{Fixed-budget and adaptive inference using the same trained SVR policy and greedy decoding. Fixed-budget inference returns the turn-$K$ answer for every example, whereas the highlighted markers report adaptive inference at its observed mean number of executed turns.}
\label{fig:fixed_adaptive}
\vspace{-5pt}
\end{figure}

The fixed-budget curves in Figure~\ref{fig:fixed_adaptive} expose the limits of uniformly increasing refinement: a larger shared budget does not consistently improve accuracy. The best fixed-budget operating points reach only 0.450 on All-7 and 0.336 on Math-5, and the preferred turn differs substantially across benchmarks. Adaptive SVR instead reaches 0.563 and 0.458 while using 2.99 and 3.42 turns on average, respectively. This advantage is not explained by generating more turns, since SVR uses fewer than four turns on average and still exceeds every evaluated fixed-budget point. Rather, the curves identify an answer-retention problem: later refinement may correct one trajectory while overwriting a correct answer in another. On GSM8K, for example, any-turn accuracy rises to 0.925 over ten turns, yet the answer returned at turn ten achieves only 0.736. The model is therefore capable of discovering correct solutions during refinement, but no common turn index can preserve them consistently. SVR addresses this mismatch by using its own self-verification to retain different turn-level answers for different inputs.

We next examine the sensitivity of adaptive inference to the confidence threshold. A useful stopping signal should remain effective without benchmark-specific tuning, so we sweep $\gamma\in\{0.50,0.55,\ldots,\\0.95\}$ while holding $T_{\max}=10$ fixed and evaluate both final-answer accuracy and PSE. The same global threshold $\gamma=0.85$ is used throughout the main experiments; the sweep is diagnostic and no dataset-specific optimum is substituted into the reported results.

\begin{figure}[h]
\centering
\includegraphics[width=\columnwidth]{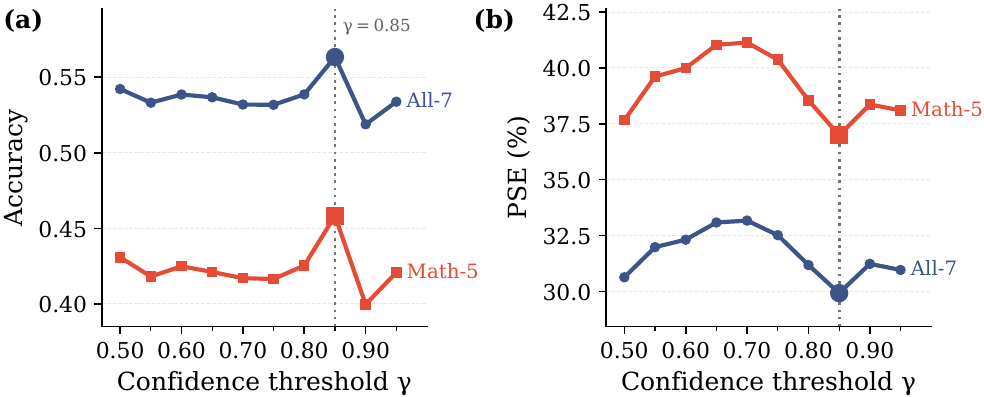}
\vspace{-11pt}
\caption{Sensitivity of adaptive SVR to the confidence threshold $\gamma$. Panel (a) reports final-answer accuracy and panel (b) reports PSE on All-7 and Math-5 with $T_{\max}=10$. The dotted line marks the shared threshold $\gamma=0.85$ used in the main experiments.}
\label{fig:threshold_sensitivity}
\vspace{-11pt}
\end{figure}

The threshold sweep in Figure~\ref{fig:threshold_sensitivity} is distinctly non-monotonic in stopping error. Within the evaluated grid, $\gamma=0.85$ provides the strongest aggregate operating point, attaining the highest All-7 and Math-5 accuracy together with the lowest aggregate PSE. Lower thresholds can allow an incorrect high-confidence assessment to terminate refinement prematurely, whereas higher thresholds may postpone termination beyond a correct intermediate answer and expose it to a later regression. This latter failure mode is consistent with the non-monotonic fixed-budget curves in Figure~\ref{fig:fixed_adaptive}. At the shared threshold, SVR stops before exhausting the budget on 86.3\% of All-7 examples and 82.8\% of Math-5 examples, confirming that the reduction in mean turns comes from active instance-dependent stopping. At the same time, PSE remains 29.9\% and 37.0\%, indicating that erroneous early commitment is still a meaningful limitation, particularly on difficult mathematical benchmarks. Taken together, the two analyses show that self-verification provides a useful mechanism for retaining correct intermediate answers and allocating refinement selectively. We next compare this adaptive allocation strategy with majority voting to determine whether the same accuracy can be obtained by uniformly increasing the number of independent single-turn samples, before examining the contributions of the individual reward components.

Finally, we compare adaptive SVR with majority voting over $k\in\{3,5,10\}$ independent GRPO samples. Table~\ref{tab:majority_vote} contrasts this uniform sampling strategy with SVR's single history-conditioned trajectory, whose length is selected by the verdict--confidence controller. Sampling and answer-normalization details follow the protocol in Section~\ref{sec:experimental_setup}.

\begin{table}[h]
\centering
\small
\setlength{\tabcolsep}{3pt}
\renewcommand{\arraystretch}{1.08}
\caption{Comparison with GRPO majority voting. Maj@$k$ uses $k$ independent samples; SVR uses adaptive inference with $\gamma=0.85$ and $T_{\max}=10$.}
\label{tab:majority_vote}
\vspace{-11pt}
\begin{tabularx}{\columnwidth}{@{}l*{4}{>{\centering\arraybackslash}X}@{}}
\toprule
\textbf{Method}
& \textbf{\shortstack{All-7 Acc.}}
& \textbf{\shortstack{All-7 Tok.}}
& \textbf{\shortstack{Math-5 Acc.}}
& \textbf{\shortstack{Math-5 Tok.}} \\
\midrule
GRPO Maj@3
& 0.504
& 5.36
& 0.404
& 6.76 \\
GRPO Maj@5
& 0.529
& 8.80
& 0.420
& 11.13 \\
GRPO Maj@10
& 0.564
& 17.50
& 0.460
& 22.15 \\
\midrule
\textbf{SVR}
& 0.563
& 8.56
& 0.458
& 11.09 \\
\bottomrule
\end{tabularx}
\vspace{-11pt}
\end{table}

At nearly matched token budgets, SVR outperforms GRPO Maj@5 by $3.4$ points on All-7 ($0.563$ vs.\ $0.529$) and $3.8$ points on Math-5 ($0.458$ vs.\ $0.420$). Maj@10 reaches comparable accuracy, differing from SVR by only $0.1$ and $0.2$ points, but consumes approximately twice as many tokens on both aggregates. Thus, SVR attains the accuracy of ten-sample voting with approximately the computation of five-sample voting. The comparison indicates that history-conditioned refinement with instance-dependent stopping provides a more favorable accuracy--compute trade-off than allocating the same number of independent samples to every input.

\subsection{Ablation Studies}

We ablate two aspects of SVR: the reward components that shape self-verification and the structure of the refinement controller. Unless otherwise specified, all adaptive variants use the same backbone, training data, optimization configuration, and inference protocol with $\gamma=0.85$ and $T_{\max}=10$. In Table~\ref{tab:reward_ablation}, $R_{\mathrm{verify}}$ denotes the complete self-verification reward block comprising calibration, asymmetric overconfidence control, error detection, and stop readiness. We independently remove the calibration, overconfidence, and stop-readiness components; removing the complete $R_{\mathrm{verify}}$ block additionally removes error detection.

\begin{table}[t]
\centering
\scriptsize
\renewcommand{\arraystretch}{1.08}
\caption{Reward-component ablations under adaptive inference. Accuracy is reported on All-7 and Math-5. Turns, PSE, Brier, and Overconf. are All-7 macro-averages; PSE and Overconf. are percentages. Diagnostic metrics should be interpreted jointly with accuracy and computation.}
\label{tab:reward_ablation}
\vspace{-11pt}
\begin{tabular*}{\columnwidth}{@{\extracolsep{\fill}}lcccccc@{}}
\toprule
Variant
& All-7
& Math-5
& Turns
& PSE
& Brier
& Overconf. \\
\midrule
Full SVR
& 0.563
& 0.458
& 2.99
& 29.9
& 0.271
& 19.3 \\
w/o $R_{\mathrm{verify}}$
& 0.528
& 0.415
& 3.13
& 30.2
& 0.288
& 21.4 \\
w/o $R_{\mathrm{cal}}$
& 0.536
& 0.432
& 3.11
& 33.8
& 0.289
& 21.2 \\
w/o $R_{\mathrm{over}}$
& 0.518
& 0.425
& 3.07
& 33.8
& 0.301
& 22.5 \\
w/o $R_{\mathrm{ready}}$
& 0.494
& 0.443
& 3.83
& 28.6
& 0.293
& 20.2 \\
\bottomrule
\end{tabular*}
\end{table}

Table~\ref{tab:reward_ablation} shows that the complete reward design provides the strongest joint operating point across task accuracy, computation, and self-verification quality. Removing the entire $R_{\mathrm{verify}}$ block reduces All-7 and Math-5 accuracy by 3.5 and 4.3 percentage points, respectively, while worsening both Brier score and overconfidence. Because this variant retains the structured output interface, solve-related objective, and adaptive stopping rule, the degradation indicates that solution shaping alone does not produce a self-verification signal with comparable utility for answer retention and compute control.

The calibration and asymmetric overconfidence components address related but distinct failure modes. Removing $R_{\mathrm{over}}$ raises PSE from 29.9\% to 33.8\% and Overconf. from 19.3\% to 22.5\%, consistent with its role in suppressing incorrect answers that are nevertheless accompanied by a positive commitment. Removing $R_{\mathrm{cal}}$ produces the same increase in PSE and worsens the Brier score from 0.271 to 0.289. These results support complementary interpretations: Brier-style calibration broadly aligns numerical confidence with correctness, whereas asymmetric overconfidence control concentrates supervision on incorrect commitments that are particularly hazardous for adaptive stopping.

The stop-readiness component determines whether a correct state becomes actionable under the stopping gate rather than directly minimizing executed computation during fixed-horizon training. Removing $R_{\mathrm{ready}}$ increases the mean inference cost from 2.99 to 3.83 turns and decreases All-7 accuracy by 6.9 percentage points. Its slightly lower PSE does not indicate a more reliable controller: the variant stops less readily, thereby reducing its opportunities to make premature stopping errors, but the additional refinement does not recover the lost task performance. PSE must therefore be interpreted jointly with final accuracy and computation.

\begin{table}[t]
\centering
\footnotesize
\renewcommand{\arraystretch}{1.08}
\caption{Structural and controller-interface ablations. Accuracy is reported on All-7 and Math-5. Turns, ESR, and PSE are All-7 macro-averages; ESR and PSE are percentages. The single-turn reference has no adaptive multi-turn stopping decision.}
\vspace{-11pt}
\label{tab:controller_ablation}
\begin{tabular*}{\columnwidth}{@{\extracolsep{\fill}}lccccc@{}}
\toprule
Variant
& All-7
& Math-5
& Turns
& ESR
& PSE \\
\midrule
Full SVR
& 0.563
& 0.458
& 2.99
& 86.3
& 29.9 \\
Single-turn SVR
& 0.471
& 0.375
& 1.00
& --
& -- \\
Verdict-only
& 0.529
& 0.438
& 2.99
& 85.2
& 32.3 \\
Confidence-only
& 0.482
& 0.384
& 5.99
& 57.8
& 20.4 \\
\bottomrule
\end{tabular*}
\end{table}

Table~\ref{tab:controller_ablation} separates the contribution of iterative refinement from that of the joint verdict--confidence interface. Single-turn SVR retains the structured self-check and applicable one-turn reward terms but removes the opportunity for repeated correction and answer preservation. The verdict-only and confidence-only policies are independently retrained, emit only the retained self-verification field, and apply the corresponding single-signal stopping rule. The former stops on a non-truncated Correct verdict, whereas the latter stops when its confidence reaches $\gamma$.

Restricting SVR to one turn reduces All-7 and Math-5 accuracy by 9.2 and 8.3 percentage points, respectively, establishing that structured self-assessment alone does not account for the gains without iterative refinement. Verdict-only retains nearly the same aggregate turn count and stopping frequency as Full SVR but lowers accuracy to 0.529 on All-7 and 0.438 on Math-5. The verdict provides a categorical commitment, but without confidence the controller lacks a continuous notion of acceptance strength. Confidence-only incurs a larger accuracy reduction and nearly doubles the mean turn count from 2.99 to 5.99 while lowering ESR from 86.3\% to 57.8\%. Its lower PSE therefore does not indicate a superior controller, because the variant terminates substantially less often and consumes considerably more refinement. Taken together, the ablations show that iterative refinement, trajectory-level self-verification supervision, calibration, asymmetric overconfidence control, stop readiness, and the joint verdict--confidence interface play complementary roles in SVR's accuracy--compute trade-off.

\section{Conclusion}

We introduced Self-Verifying Refinement (SVR), an oracle-free multi-turn reinforcement learning framework that uses learned verdict--confidence signals to allocate test-time computation. Across seven mathematical reasoning benchmarks with Qwen3.5-2B, SVR achieves $0.563$ macro-average accuracy with $2.99$ turns and $8.56$ thousand tokens per example, scoring above the evaluated single-turn and fixed-budget multi-turn references in the complete-system comparison. It also matches ten-sample GRPO majority voting at approximately half the token cost. Analyses attribute these gains to instance-dependent stopping and complementary self-verification signals, while reducing premature-stop errors on difficult problems remains an important direction.

\bibliographystyle{ACM-Reference-Format}
\bibliography{sample-base}


\appendix

\section{Implementation Details}

\subsection{Training and Generation Hyperparameters}
\label{app:training_hyperparameters}

All trainable methods use full-parameter reinforcement learning on the post-trained \path{Qwen/Qwen3.5-2B} checkpoint within the ms-swift framework. Training is conducted in bfloat16 precision with DeepSpeed ZeRO-2 and gradient checkpointing. Rollout generation is colocated with training through vLLM using a GPU-memory utilization ratio of $0.4$, tensor-parallel size $1$, and a maximum model context length of $8192$ tokens. All experiments are executed on four NVIDIA A800 GPUs. The infrastructure is shared across methods whenever applicable, while the policy objective, multi-turn scheduler, prompt construction, and reward function follow the corresponding method definitions.

All methods are trained for one epoch using fused AdamW with a learning rate of $4\times10^{-8}$, weight decay $0.1$, $(\beta_1,\beta_2)=(0.9,0.95)$, and gradient clipping at $2.0$. The learning-rate schedule follows Warmup--Stable--Decay, with a $5\%$ linear warmup phase and a $20\%$ linear decay tail that terminates at $5\%$ of the peak learning rate. The per-device training batch size is $8$. Multi-turn methods with $G=8$ trajectories use gradient accumulation $4$, whereas single-turn GRPO and GSPO with $G=24$ completions use gradient accumulation $12$. Both settings process $16$ distinct input problems per optimizer step.

Training rollouts are sampled with temperature $0.9$ and top-$p=1.0$. GRPO-based objectives use clipping coefficient $\epsilon=0.2$. Unless otherwise specified, the KL coefficient is initialized at $\beta=0.08$, and the reference policy is synchronized with mixing coefficient $0.6$ every $80$ optimizer steps on Countdown and every $40$ steps on GSM8K and MATH. Rewards are not additionally rescaled, and dynamic resampling permits at most three generation attempts. The primary runs use random seed $42$; experiments evaluated over additional training seeds change only the seed unless explicitly stated otherwise.

Countdown uses $50{,}000$ training examples sampled from \path{Jiayi-Pan/Countdown-Tasks-3to4} with seed $42$, while GSM8K and MATH use their complete training splits of $7{,}473$ and $7{,}500$ examples, respectively. With an effective batch of $16$ input problems per optimizer step, one epoch corresponds to approximately $3{,}125$, $468$, and $469$ optimizer steps for Countdown, GSM8K, and MATH, respectively. Completion and length-control budgets are selected according to the typical reasoning length of each domain, as summarized in Table~\ref{tab:task_generation_budgets}.

\begin{table}[h]
\centering
\small
\setlength{\tabcolsep}{4pt}
\caption{Task-dependent generation and length-control parameters used during training. The length-normalization scale $L_{\mathrm{len}}$ is used by the solve-related reward. The long-output thresholds apply only to methods for which the corresponding auxiliary safeguard is enabled.}
\label{tab:task_generation_budgets}
\begin{tabular}{lccc}
\toprule
\textbf{Hyperparameter}
& \textbf{Countdown}
& \textbf{GSM8K}
& \textbf{MATH} \\
\midrule
Maximum completion tokens $L_{\mathrm{gen}}$
& 800 & 1200 & 2048 \\
Length-normalization scale $L_{\mathrm{len}}$
& 768 & 1024 & 1536 \\
Long-output expected length
& 600 & 900 & 1400 \\
Long-output cache length
& 200 & 300 & 400 \\
\bottomrule
\end{tabular}
\end{table}

Prompt-side context is truncated to at most $4096$ tokens, while the complete prompt--completion sequence remains bounded by the $8192$-token model context. Previous responses included in refinement prompts are truncated to $512$, $640$, and $900$ characters for Countdown, GSM8K, and MATH, respectively. When this limit is exceeded, a truncation marker is appended so that the next-turn policy can distinguish a shortened draft from a complete previous response.

SVR and the oracle-guided score-feedback reference use a fixed training horizon of $T_{\mathrm{tr}}=3$ with $G=8$ sampled trajectories per input, yielding $3\times8=24$ turn-level generations. Adaptive stopping is disabled during training, so every sampled trajectory executes all three turns. For SVR, however, these generations produce only eight optimization samples: solve, self-verification, and format quantities are computed at all three turns, averaged into one scalar per trajectory, and the policy-gradient loss covers only the third-turn completion. The first two turns provide context for subsequent prompts and contribute to the trajectory return, but are not independently optimized. GRPO-MT, Murphy, MLMT-RL, and ScRPO use the same three-turn sampling structure and generation count. Single-turn GRPO and GSPO instead sample $G=24$ independent completions per input; Countdown GRPO and GSPO use a maximum completion length of $2048$ tokens, while their remaining task-dependent generation limits follow Table~\ref{tab:task_generation_budgets}. iGRPO uses a two-stage schedule in which the first stage samples $12$ exploratory drafts and the second stage samples $12$ refinements conditioned on the highest-reward draft, with policy-gradient updates applied only to the second-stage outputs. The task-specific correctness and format evaluators are shared across applicable methods, while SVR additionally evaluates its structured self-check. Exact reward definitions and coefficients are reported in Section~\ref{app:reward-details}.

All reported evaluations use greedy decoding with temperature $0$. Adaptive SVR stops at the first non-truncated turn satisfying $v_t=\mathrm{C}$ and $c_t\geq0.85$, subject to a maximum inference budget of $T_{\max}=10$. The maximum completion lengths used at evaluation are $800$ tokens for Countdown, $1200$ for GSM8K, $2048$ for MATH500 and AMC23, $3072$ for AIME26 and MinervaMath, and $4096$ for OlympiadBench. Fixed-budget multi-turn evaluations use the same decoding and task-dependent completion limits but disable confidence-gated stopping and return the answer generated at the prescribed final turn.

\subsection{Reward Specification and Coefficients}
\label{app:reward-details}

This section provides the complete reward specification underlying Section~\ref{sec:learning-controller}. Unless otherwise stated, the coefficients are shared across Countdown, GSM8K, and MATH. For each fixed-horizon trajectory \(\tau=(o_1,\ldots,o_{T_{\mathrm{tr}}})\), the per-turn solve, self-verification, and format signals are averaged into
\begin{equation}
R_{\mathrm{SVR}}(\tau)
=
\frac{1}{T_{\mathrm{tr}}}
\sum_{t=1}^{T_{\mathrm{tr}}}
\left(
r_{\mathrm{solve},t}
+
r_{\mathrm{verify},t}
+
\lambda_{\mathrm{fmt}}r_{\mathrm{fmt},t}
\right).
\label{eq:app-svr-reward}
\end{equation}
The evaluator assigns \(y_t=1\) when the extracted task answer is correct and \(y_t=0\) otherwise; a missing or unparseable answer is therefore treated as incorrect. The truncation indicator \(q_t=1\) records that generation reached the completion-token limit. Metrics used only for training diagnostics have zero trainer weight and do not contribute to Eq.~\eqref{eq:app-svr-reward}.

For the solve component, let \(\ell_t\) denote the number of generated completion tokens and \(L_{\mathrm{len}}\) the task-dependent length-normalization scale reported in the preceding subsection. We define the length-shaped correctness and absolute reward as
\begin{equation}
\widetilde{y}_t
=
y_t
\left(
1-
\alpha
\frac{\min(\ell_t,L_{\mathrm{len}})}
{L_{\mathrm{len}}}
\right),
\qquad
r_{\mathrm{abs},t}
=
\lambda_{\mathrm{abs}}\widetilde{y}_t.
\label{eq:app-absolute-reward}
\end{equation}
Because \(\widetilde{y}_t=0\) whenever \(y_t=0\), shorter incorrect answers receive no positive length reward. The cross-turn progress component is
\begin{equation}
r_{\mathrm{prog},t}
=
\begin{cases}
0, & t=1,\\
\lambda_{\Delta}(y_t-y_{t-1})
+\phi(y_{t-1},y_t), & t\geq2,
\end{cases}
\label{eq:app-progress-reward}
\end{equation}
where
\begin{equation}
\phi(y_{t-1},y_t)
=
\begin{cases}
\lambda_{\mathrm{keep}}, & (y_{t-1},y_t)=(1,1),\\
-\lambda_{\mathrm{reg}}, & (y_{t-1},y_t)=(1,0),\\
-\lambda_{\mathrm{fail}}, & (y_{t-1},y_t)=(0,0),\\
0, & (y_{t-1},y_t)=(0,1).
\end{cases}
\label{eq:app-transition-cases}
\end{equation}
The improvement term \(\lambda_{\Delta}(y_t-y_{t-1})\) rewards wrong-to-correct transitions and penalizes correct-to-wrong transitions, while \(\phi\) separately rewards preservation and penalizes regression or repeated failure. The complete per-turn solve reward is
\begin{equation}
r_{\mathrm{solve},t}
=
r_{\mathrm{abs},t}
+
r_{\mathrm{prog},t}
-
\lambda_{\mathrm{trunc}}q_t.
\label{eq:app-solve-reward}
\end{equation}
A truncated completion is still evaluated according to any task answer that can be extracted from its realized text, but it receives the additional truncation penalty in Eq.~\eqref{eq:app-solve-reward}.

The self-verification reward is applied only to complete generations:
\begin{equation}
\begin{aligned}
r_{\mathrm{verify},t}
&=
(1-q_t)
\Bigl(
\lambda_{\mathrm{cal}}\left[1-(c_t-y_t)^2\right]
-\lambda_{\mathrm{over}}c_t
\mathbb{I}[v_t=\mathrm{C}\land y_t=0]\\
&+\lambda_{\mathrm{detect}}
\mathbb{I}[v_t=\mathrm{I}\land y_t=0]
+\lambda_{\mathrm{ready}}c_t
\mathbb{I}[v_t=\mathrm{C}\land y_t=1]
\Bigr).
\end{aligned}
\label{eq:app-verification-reward}
\end{equation}
The multiplicative mask sets the entire self-verification reward to zero when \(q_t=1\), since the verdict or confidence may be incomplete. For non-truncated outputs, the four terms respectively implement Brier-style confidence calibration, asymmetric penalization of confidently incorrect \textsc{Correct} judgments, explicit error detection, and stop readiness for correctly solved states. The deployment threshold \(\gamma\) is not used in Eq.~\eqref{eq:app-verification-reward} or in training-time trajectory termination.

At every turn, the format component combines independent task-answer and self-check scores:
\begin{equation}
r_{\mathrm{fmt},t}
=
r_{\mathrm{taskfmt},t}
+
r_{\mathrm{scfmt},t},
\label{eq:app-format-reward}
\end{equation}
where the task-specific \(r_{\mathrm{taskfmt},t}\) evaluates whether the answer follows the required Countdown, GSM8K, or MATH output convention. Its exact domain-specific rules are provided with the answer evaluators in Appendix~C.2. The self-check score is
\begin{equation}
r_{\mathrm{scfmt},t}
=
\begin{cases}
1, & \text{a complete self-check block contains a parseable}
\\ &\text{verdict and confidence},\\
0, & \text{the self-check block is present but malformed},\\
-1, & \text{the self-check block is absent}.
\end{cases}
\label{eq:app-self-check-format}
\end{equation}
Unlike \(r_{\mathrm{verify},t}\), the realized format scores are retained when \(q_t=1\), so a truncated output is still evaluated according to the structure actually generated. Tolerant controller parsing is independent of \(r_{\mathrm{scfmt},t}\): fields recovered outside a complete canonical block may be used for refinement, while the format reward continues to penalize violation of the required serialization.

\begin{table}[t]
\centering
\small
\setlength{\tabcolsep}{3pt}
\renewcommand{\arraystretch}{1.08}
\caption{Reward coefficients used for SVR training.}
\label{tab:reward-coefficients}
\begin{tabular*}{\columnwidth}{@{\extracolsep{\fill}}lp{0.62\columnwidth}c}
\toprule
\textbf{Symbol} & \textbf{Role} & \textbf{Value} \\
\midrule
\(\lambda_{\mathrm{abs}}\) & Absolute correctness reward & 1.0 \\
\(\lambda_{\Delta}\) & Turn-to-turn improvement weight & 0.3 \\
\(\alpha\) & Completion-length shaping strength & 0.5 \\
\(\lambda_{\mathrm{keep}}\) & Correct-answer preservation bonus & 0.3 \\
\(\lambda_{\mathrm{reg}}\) & Correct-to-incorrect regression penalty & 0.5 \\
\(\lambda_{\mathrm{fail}}\) & Repeated-failure penalty & 0.3 \\
\(\lambda_{\mathrm{trunc}}\) & Generation-truncation penalty & 0.5 \\
\(\lambda_{\mathrm{cal}}\) & Brier-style calibration reward & 0.5 \\
\(\lambda_{\mathrm{over}}\) & Asymmetric overconfidence penalty & 0.8 \\
\(\lambda_{\mathrm{detect}}\) & Explicit error-detection reward & 0.2 \\
\(\lambda_{\mathrm{ready}}\) & Stop-readiness reward & 0.3 \\
\(\lambda_{\mathrm{fmt}}\) & Task-answer and self-check format weight & 0.4 \\
\bottomrule
\end{tabular*}
\end{table}

After all \(T_{\mathrm{tr}}\) turns have been generated, Eq.~\eqref{eq:app-svr-reward} produces one scalar return for each trajectory. This return supervises all generated tokens of the final completion only. Earlier completions contribute to subsequent refinement contexts and to the trajectory-averaged reward, but remain outside the policy-gradient loss and are not treated as independent GRPO samples.

The oracle-guided score-feedback reference uses the same solve-related coefficients and task-answer format weight, but it does not optimize \(r_{\mathrm{verify},t}\) or the self-check-format component. Other baselines retain their method-specific reward and credit-assignment rules described in the experimental setup.

\section{Prompt Templates and Output Parsing}

\subsection{SVR Prompt Templates}
\label{app:svr_prompt_templates}

SVR uses a Markov prompt schedule in which every generation receives only the system instruction and one user message. At the first turn, the user message contains the original problem alone. At each subsequent turn, the dialogue history is reset and the new user message is constructed from the original problem, a length-bounded copy of the immediately preceding completion, and the verdict--confidence pair parsed from that completion. Ground-truth correctness labels, reference answers, reward values, and external verifier signals are never inserted into an SVR prompt. Training and inference use the same system instruction, output interface, and refinement templates; they differ only in that training executes the complete horizon $T_{\mathrm{tr}}=3$, whereas inference may terminate before $T_{\max}$ when the confidence gate is satisfied.

All SVR policies use the same role instruction with a task-specific answer format. For GSM8K and MATH, the system message is:

\begin{quote}
\small\ttfamily
You are a careful math assistant who audits your own work.\\
First reason step-by-step inside <think>...</think>. Then give the final answer inside <answer>\textbackslash boxed\{YOUR\_\\
ANSWER\}</answer>.\\
Finally, on a new line, output a self-verification in EXACTLY this form: <self\_check>VERDICT: CORRECT, INCORRECT, or UNSURE; CONFIDENCE: a number in [0,1] </self\_check> \\
Be honest and well-calibrated: only report CORRECT with high confidence after verifying every step. Always use \textbackslash boxed\{\} for the final answer.
\end{quote}

For Countdown, the answer instruction is changed to \texttt{<answer>\\YOUR\_EQUATION</answer>}, and the final requirement to use \texttt{\textbackslash boxed\{\}} is omitted. Thus, every response contains a reasoning trace, a task answer, and a structured self-check in the following order:

\begin{quote}
\small\ttfamily
<think>... step-by-step derivation ...</think>\\
<answer>\textbackslash boxed\{42\}</answer>\\
<self\_check>VERDICT: CORRECT; CONFIDENCE: 0.92\\</self\_check>
\end{quote}

The boxed answer in this example applies to GSM8K and MATH. Countdown instead requires a plain executable equation inside the answer block. The structured self-check exposes the discrete verdict $v_t\in\{\mathrm{C},\mathrm{I},\mathrm{U}\}$ and numerical confidence $c_t\in[0,1]$ used by the refinement controller. Its parsing and fallback behavior are specified in Section~\ref{app:self_check_parsing}.

At turn $t=1$, the user message contains only the raw input problem:

\begin{quote}
\small\ttfamily
\{original question\}
\end{quote}

No turn identifier, previous response, correctness signal, or verifier feedback is included. This keeps the initial policy interface identical to ordinary single-turn generation and attributes subsequent changes to the refinement process.

For each turn $t\geq2$, let $(v_{t-1},c_{t-1})$ denote the self-verification parsed from the preceding completion, and let $d_{t-1}$ denote that completion after prompt-side truncation. The draft limits are $512$, $640$, and $900$ characters for Countdown, GSM8K, and MATH, respectively. When a completion exceeds the corresponding character limit, the retained prefix is followed by \texttt{[...truncated]}. This prompt-side shortening is distinct from generation truncation: the \texttt{TRUNCATED} branch below is activated only when the preceding generation itself reached its token limit.

The turn-$t$ user message follows the template:

\begin{quote}
\small\ttfamily
[T=\{t\}] Your self-verification last turn: \{header\}\\
Question: \{original question\}\\[2pt]
Your previous solution:\\
\{previous completion\}\\[2pt]
\{refinement instruction\}
\end{quote}

For a complete preceding response, \texttt{\{header\}} is formatted as \texttt{\{VERDICT\} (conf \{CONFIDENCE\})}, with confidence rounded to two decimal places. If the generation reached its token limit, the header is set to \texttt{TRUNCATED}. The refinement instruction is selected in the priority order generation truncation, verdict \textsc{Correct}, and otherwise.

When the preceding response was cut off by the generation-length limit, its partial answer and self-verification are treated as unreliable. SVR uses the following regeneration instruction:

\begin{quote}
\small\ttfamily
Your previous response was CUT OFF before completion.\\
Discard it and produce a fresh, COMPLETE solution. Keep\\
<think>...</think> concise so the entire answer (including\\
<answer>\textbackslash boxed\{\}</answer> and the <self\_check> block) fits\\
within the budget.
\end{quote}

When $v_{t-1}=\mathrm{C}$ and another turn is required, the previous answer is not treated as externally verified. Instead, the model is instructed to independently audit its most error-prone step before preserving or revising the solution:

\begin{quote}
\small\ttfamily
In your previous attempt you judged the answer CORRECT\\
(self-confidence \{c\}), with NO external confirmation.\\
Independently re-derive the single most error-prone step.\\
If it still holds, restate the SAME final answer in\\
<answer>\textbackslash boxed\{\}</answer> and report VERDICT: CORRECT.\\
If you now find a mistake, fix it and report your updated\\
verdict honestly.
\end{quote}

When $v_{t-1}\in\{\mathrm{I},\mathrm{U}\}$, the next prompt requests explicit error localization and a complete revision:

\begin{quote}
\small\ttfamily
In your previous attempt you judged the answer likely WRONG\\
or were unsure (self-confidence \{c\}). Locate the specific\\
logical or arithmetic error, then produce a corrected,\\
complete step-by-step solution with the final answer in\\
<answer>\textbackslash boxed\{\}</answer> and an honest <self\_check> block.
\end{quote}

The three refinement instructions above show the GSM8K and MATH answer syntax. For Countdown, every occurrence of \texttt{<answer>\\\textbackslash boxed\{\}</answer>} is replaced by a plain equation inside \texttt{<answer>\\...</answer>}. Apart from this task-specific answer representation, the prompt transition is shared across the three training domains.

This construction preserves the oracle-free information boundary throughout the refinement trajectory. Every subsequent prompt depends only on the original problem and policy-generated information from the immediately preceding response. The task evaluator may compute the binary correctness label $y_t$ for rewards and diagnostic metrics, but neither $y_t$ nor any derived score affects the SVR prompt. Consequently, the refinement policy receives the same type of information during training and deployment, while its own verdict and confidence jointly determine the refinement instruction and, at inference time, whether additional computation is allocated.

\subsection{Oracle-Guided Baseline Prompt}
\label{app:oracle_guided_prompt}

The oracle-guided score-feedback reference uses the same Markov context structure and task-specific answer interface as SVR, but differs in the information exposed to the refinement policy. After each complete generation, the task evaluator computes a binary correctness label $y_{t-1}\in\{0,1\}$ for the preceding answer. This label is inserted into the next user message and directly determines the refinement instruction. The reference therefore receives privileged correctness feedback during refinement, whereas SVR constructs every subsequent prompt exclusively from policy-generated information.

The oracle-guided reference does not generate a structured self-check. For Countdown, the response contains a reasoning trace followed by an executable equation inside \texttt{<answer>...</answer>}. For GSM8K and MATH, the final result is enclosed in \texttt{<answer>\textbackslash \\boxed\{...\}</answer>}. No verdict or confidence field is requested, parsed, or used for stopping. Consequently, training executes the complete horizon $T_{\mathrm{tr}}=3$, and the main evaluation uses the full fixed inference budget rather than confidence-gated adaptive termination.

At turn $t=1$, the user message contains only the original problem. For each turn $t\geq2$, the dialogue history is reset to the system message and a newly constructed user message. Let $d_{t-1}$ denote the preceding completion after prompt-side truncation. The draft limits are $512$, $640$, and $900$ characters for Countdown, GSM8K, and MATH, respectively. If the retained draft exceeds the corresponding limit, its prefix is followed by \texttt{[...truncated]}. As in SVR, this prompt-side shortening is distinct from generation truncation caused by reaching the completion-token limit.

For a complete preceding response, the turn-$t$ prompt is:

\begin{quote}
\small\ttfamily
[T=\{t\}] Score: \{$y_{t-1}$ formatted to three decimals\}\\
Question: \{original question\}\\[2pt]
Your previous solution:\\
\{previous completion\}\\[2pt]
\{score-conditioned refinement instruction\}
\end{quote}

Because the evaluators used in all reported experiments return binary labels, the displayed score is either \texttt{0.000} or \texttt{1.000}. No intermediate or partially correct score is used by the reported oracle-guided runs.

When $y_{t-1}=1$, the evaluator has marked the preceding answer as correct, and the following instruction is used:

\begin{quote}
\small\ttfamily
Your previous answer is CORRECT. Briefly verify the reasoning,\\
then output the SAME final answer again inside\\
<answer>\textbackslash boxed\{\}</answer>.
\end{quote}

When $y_{t-1}=0$, the evaluator has marked the preceding answer as incorrect, and the model receives an explicit correction request:

\begin{quote}
\small\ttfamily
Your previous answer is INCORRECT. Carefully review the\\
previous solution, identify the logical or arithmetic errors,\\
and provide a correct step-by-step solution.
\end{quote}

If the preceding generation reached its token limit, the evaluator label is not used to characterize the incomplete response. The score header is instead set to \texttt{TRUNCATED}, and the following regeneration instruction takes priority:

\begin{quote}
\small\ttfamily
Your previous response was CUT OFF before completion\\
(it hit the length budget mid-stream). The draft above is\\
incomplete and should not be trusted. Discard it, and produce\\
a fresh, COMPLETE solution this time. Keep your\\
<think>...</think> reasoning concise so the entire answer\\
including <answer>...</answer> fits within the budget.
\end{quote}

The templates above show the GSM8K and MATH answer syntax. For Countdown, every occurrence of \texttt{<answer>\textbackslash boxed\{\}</answer>} is replaced by a plain executable equation inside \texttt{<answer>...\\</answer>}. Apart from this task-specific output representation, the prompt construction and correctness-conditioned branches are shared across the three training domains.

This reference isolates a practically strong form of multi-turn refinement with policy-visible evaluator feedback, but it is not an identical-controller ablation of SVR. Its prompts contain privileged correctness information, its output interface contains no policy-generated self-verification, and it provides no internal signal for adaptive stopping. It is therefore evaluated as a fixed-budget score-feedback reference rather than as an oracle-assisted version of the SVR controller.

\subsection{Self-Check Parsing and Fallback Rules}
\label{app:self_check_parsing}

SVR converts each generated self-check into a discrete verdict $v_t\in\{\mathrm{C},\mathrm{I},\mathrm{U}\}$ and a confidence value $c_t\in[0,1]$. Parsing is case-insensitive and deliberately tolerates minor surface-form deviations so that the refinement controller does not fail solely because of capitalization, separators, or a missing wrapper tag. The parser first searches for the first complete \texttt{<self\_check>...\allowbreak</self\_check>} span, allowing its contents to extend across multiple lines. If such a span is found, both fields are extracted only from within that span. If no complete span is present, the parser scans the full completion for explicit \texttt{VERDICT} and \texttt{CONFIDENCE} fields as a robustness fallback.

The accepted verdict strings are \texttt{CORRECT}, \texttt{INCORRECT}, and \texttt{UNSURE}. The aliases \texttt{RIGHT} and \texttt{WRONG} are normalized to \texttt{CORRECT} and \texttt{IN-\\CORRECT}, respectively. Verdict names may be preceded by either a colon, an equals sign, or whitespace. Confidence may be written as a decimal, such as \texttt{0.92}, or in percentage form, such as \texttt{92\%}. A parseable numerical value greater than $1$ is interpreted as percentage-style and divided by $100$, after which the result is clipped to $[0,1]$.

Missing or malformed fields are resolved independently. An unparseable verdict is mapped to \textsc{Unsure} while retaining any confidence that can still be extracted. If confidence is unavailable, SVR assigns a verdict-dependent prior:
\begin{equation}
c_t=
\begin{cases}
0.8, & v_t=\mathrm{C},\\
0.2, & v_t=\mathrm{I},\\
0.5, & v_t=\mathrm{U}.
\end{cases}
\label{eq:confidence_fallback}
\end{equation}
An empty completion or a response from which neither field can be recovered therefore yields $(v_t,c_t)=(\mathrm{U},0.5)$. These defaults are conservative with respect to the deployment threshold $\gamma=0.85$: a missing verdict cannot satisfy the discrete stopping condition, and a \textsc{Correct} verdict without an explicit confidence receives $0.8<\gamma$ and therefore cannot trigger early termination.

\begin{table}[t]
\centering
\scriptsize
\setlength{\tabcolsep}{4pt}
\caption{Parsing and fallback rules for the structured self-check.}
\label{tab:self_check_parsing}
\begin{tabularx}{\columnwidth}{@{}p{0.29\columnwidth}X@{}}
\toprule
\textbf{Condition} & \textbf{Parsing rule} \\
\midrule
Complete self-check block & Restrict verdict and confidence extraction to the first complete \texttt{<self\_check>...</self\_check>} span. \\
Missing complete block & Scan the full completion for explicit \texttt{VERDICT} and \texttt{CONFIDENCE} fields. \\
Verdict field & Accept \texttt{CORRECT}, \texttt{INCORRECT}, and \texttt{UNSURE}; normalize \texttt{RIGHT} and \texttt{WRONG}. \\
Missing or invalid verdict & Set $v_t=\mathrm{U}$ while retaining any parseable confidence. \\
Confidence field & Accept decimal and percentage forms; divide percentage-style values by $100$ and clip to $[0,1]$. \\
Missing confidence & Use $0.8$, $0.2$, or $0.5$ for \textsc{Correct}, \textsc{Incorrect}, or \textsc{Unsure}, respectively. \\
Empty or unparseable output & Return $(v_t,c_t)=(\mathrm{U},0.5)$. \\
Generation truncation & Mark the response as \texttt{TRUNCATED}; it is ineligible for confidence-gated stopping. \\
\bottomrule
\end{tabularx}
\end{table}

Accepted self-check fields and their fallback behavior are catalogued in Table~\ref{tab:self_check_parsing}. At adaptive inference, a generated response terminates refinement only when all three conditions are satisfied:
\begin{equation}
\mathrm{Stop}(t)
=
\Ind[
v_t=\mathrm{C}
\land c_t\geq\gamma
\land f_t\neq\mathrm{length}
],
\label{eq:app_adaptive_stop}
\end{equation}
where $f_t$ is the generation finish reason and $\gamma=0.85$ in the main experiments. Thus, an output that reaches the completion-token limit cannot terminate the trajectory even if a partial self-check is parsed as confident and correct. When additional budget remains, generation truncation takes priority over the parsed verdict: the subsequent prompt uses the \texttt{TRUNCATED} header and requests a fresh complete response. If the stopping condition is never satisfied, SVR returns the answer generated at turn $T_{\max}$. During training, confidence-gated stopping is disabled and every trajectory executes all $T_{\mathrm{tr}}=3$ turns, although the parsed verdicts and confidences are still used to construct subsequent prompts and compute the self-verification reward.

The tolerant parser is distinct from the self-check component of the format reward described in Section~\ref{app:reward-details}. The parser may recover explicit fields located outside a complete self-check block, whereas $r_{\mathrm{scfmt},t}$ requires the wrapper itself: a complete block containing both parseable fields receives raw reward $+1$, a present but malformed block receives $0$, and an absent block receives $-1$. This separation combines robust deployment-time interpretation with explicit training pressure toward the required output interface.

Self-check parsing is also independent of task-answer evaluation. The verdict and confidence fields determine refinement prompts, self-verification rewards, diagnostic metrics, and adaptive stopping, but they never determine whether the task answer is correct. The evaluator separately extracts the candidate answer from the \texttt{<answer>...</answer>} field and assigns the binary label $y_t\in\{0,1\}$ according to the task-specific rules in Section~\ref{app:answer_evaluation}.

\section{Datasets, Evaluators, and Metrics}

\subsection{Dataset Statistics and Splits}
\label{app:dataset_statistics}

We train three separate domain-specific policies on Countdown, GSM8K, and competition MATH. The Countdown- and GSM8K-trained policies are evaluated on held-out examples from their corresponding domains, whereas the competition-MATH policy is evaluated without further fine-tuning on MATH500 and four additional mathematical reasoning benchmarks. Within each domain, all compared methods use exactly the same training examples, evaluation sets, and task-specific correctness evaluator. Table~\ref{tab:dataset_statistics} summarizes the dataset sizes and their roles in the experiments.

\begin{table}[h]
\centering
\small
\setlength{\tabcolsep}{3pt}
\caption{Training and evaluation datasets. $N$ denotes the number of examples used in the corresponding experimental role.}
\label{tab:dataset_statistics}
\begin{tabularx}{\columnwidth}{@{}lXrX@{}}
\toprule
\textbf{Domain} & \textbf{Dataset or split} & \textbf{$N$} & \textbf{Role} \\
\midrule
Countdown & Full puzzle pool & 490,364 & Source corpus \\
Countdown & Sampled training subset & 50,000 & RL training \\
Countdown & Held-out subset & 1,000 & In-domain evaluation \\
\midrule
GSM8K & Official train split & 7,473 & RL training \\
GSM8K & Official test split & 1,319 & In-domain evaluation \\
\midrule
MATH & Competition MATH train & 7,500 & RL training \\
MATH & MATH500 & 500 & Held-out evaluation \\
MATH & AIME26 & 30 & Contest evaluation \\
MATH & AMC23 & 40 & Contest evaluation \\
MATH & OlympiadBench & 674 & Olympiad evaluation \\
MATH & MinervaMath & 272 & Cross-benchmark evaluation \\
\bottomrule
\end{tabularx}
\end{table}

The Countdown-Tasks-3to4 source corpus contains $490{,}364$ arithmetic puzzles constructed from mixed three- and four-number inputs. We sample $50{,}000$ problems for reinforcement learning using a permutation generated with seed $42$. The evaluation set contains $1{,}000$ examples sampled with seed $0$ from the complementary pool after excluding all selected training indices. The training and evaluation subsets therefore have no index overlap, and all reported Countdown results use the complete held-out subset.

For GSM8K~\citep{cobbe2021training}, we use all $7{,}473$ problems in the official training split for reinforcement learning and evaluate on all $1{,}319$ examples in the official test split. The dataset answer field is mapped to the reference-solution field required by our evaluator. We apply no additional filtering, subsampling, or resampling to either split.

For the competition MATH domain~\citep{hendrycksmath2021}, reinforcement learning uses all $7{,}500$ problems in the official training split. We do not train on its test split because MATH500~\citep{lightman2023process}, which is constructed from held-out competition-MATH problems, serves as the primary evaluation set for this policy. This protocol keeps the $500$ MATH500 questions outside the reinforcement-learning data and prevents direct train--evaluation overlap.

The same competition-MATH-trained policy is additionally evaluated without further optimization on AIME26~\citep{dekoninck2026matharena}, AMC23, OlympiadBench~\citep{he2024olympiadbench}, and MinervaMath~\citep{lewkowycz2022solving}. We use all examples in the selected evaluation splits: $30$ for AIME26, $40$ for AMC23, $674$ for OlympiadBench, and $272$ for MinervaMath. No evaluation subsampling is used for the main results. Because AIME26 and AMC23 contain only $30$ and $40$ problems, respectively, their dataset-level results and confidence diagnostics should be interpreted with appropriate caution.

All main training runs cover one epoch of the corresponding training set. Differences in training-set size across the three domains reflect the available dataset constructions: Countdown uses a fixed subset of a substantially larger generated corpus, while GSM8K and competition MATH use their complete official training splits. Optimization steps and batching details are reported in Section~\ref{app:training_hyperparameters}.

\subsection{Answer Extraction and Correctness Evaluation}
\label{app:answer_evaluation}

For every generated turn, a task-specific evaluator independently extracts the candidate task answer and compares it with the corresponding reference target. The structured self-check is never used to determine correctness: the verdict and confidence affect refinement prompts, self-verification rewards, diagnostic metrics, and adaptive stopping, whereas correctness depends only on the content extracted from the task-answer field. All evaluators used in the reported experiments return a binary label
\begin{equation}
y_t=
\begin{cases}
1, & \text{if the extracted answer is correct},\\
0, & \text{otherwise}.
\end{cases}
\label{eq:app_binary_correctness}
\end{equation}
No intermediate or partial correctness scores are used for reward computation, accuracy, calibration, or stopping analysis. Missing, malformed, or unparseable answers receive $y_t=0$.

For Countdown, the evaluator extracts the first complete \texttt{<answer>\\...\allowbreak</answer>} span. If the answer block is absent or empty, the prediction is marked incorrect. When the extracted content contains an equality, such as \texttt{3*(7+1)=24}, only the expression preceding the equality sign is evaluated. A prediction is correct only when the extracted expression contains permitted arithmetic syntax, uses exactly the multiset of numbers supplied by the problem, can be evaluated successfully, and produces the requested target value. Failure of any of these conditions yields $y_t=0$; satisfying all conditions yields $y_t=1$.

For GSM8K, the evaluator first searches the final \texttt{<answer>...\allowbreak</answer>} block. If no usable answer block is found, the final $500$ characters of the completion are searched as a robustness fallback. Within the selected scope, the last expression enclosed by \texttt{\textbackslash boxed\{...\}} is preferred; otherwise, the evaluator searches for the last numerical answer following the conventional \texttt{\#\#\#\#} marker. The extracted prediction and reference answer are converted to numerical values after removing superficial formatting such as whitespace and thousands separators. The prediction receives $y_t=1$ when its absolute difference from the reference value is below $10^{-5}$ and receives $y_t=0$ otherwise.

MATH500, AIME26, AMC23, OlympiadBench, and MinervaMath use the same mathematical answer-extraction pipeline. The evaluator first searches for the final boxed expression inside the \texttt{<answer>...\allowbreak</answer>} block and falls back to the complete response when no usable boxed answer is present in that block. Brace-aware extraction preserves nested LaTeX expressions such as \texttt{\textbackslash frac\{a\}\{b\}}. The reference answer is obtained from the benchmark's gold-answer field or, when required by the dataset representation, from the final boxed expression in the provided reference solution.

For the MATH-family benchmarks, the extracted prediction and reference answer are first normalized to remove non-semantic formatting differences. Correctness is then determined through normalized symbolic-string comparison, numerical comparison with absolute tolerance $10^{-5}$ when both expressions admit numerical interpretation, and fraction-based normalization when applicable. A prediction is marked correct when one of these supported equivalence checks succeeds. The evaluator does not assign credit for partially matching derivations, intermediate reasoning steps, or a correct method followed by an incorrect final answer.

Task-answer format is evaluated separately from task correctness at every turn. For the MATH-family tasks, an \texttt{<answer>...\allowbreak</answer>} block containing a non-empty \texttt{\textbackslash boxed\{...\}} expression receives $r_{\mathrm{taskfmt},t}=1.0$, a present answer block without such a boxed expression receives $0.3$, and an absent answer block receives $0.0$. For GSM8K, the score is $1.0$ when the answer block contains either a boxed expression or a numerical answer following the \texttt{\#\#\#\#} marker and is $0.0$ otherwise. For Countdown, the score is $1.0$ only when the entire string presented to the task-format evaluator matches \texttt{\string^<think>...</think>\string\s*<answer>...</answer>\string}; otherwise, it is $0.0$. This check imposes no boxed-answer or self-check requirement. These task-format checks do not change the binary correctness label in Eq.~\eqref{eq:app_binary_correctness}. They are combined with the separately evaluated self-check-format score at each turn and then averaged over the fixed training trajectory.

Generation truncation is also handled independently from answer equivalence. A response that reaches the completion-token limit is marked as truncated and cannot activate confidence-gated stopping, even when a candidate answer or self-check can be parsed from its incomplete text. During fixed-horizon training, its task answer is still processed by the corresponding evaluator, while the solve reward additionally applies the truncation penalty described in Section~\ref{app:reward-details}. If another refinement turn remains, the next prompt uses the regeneration branch described in Section~\ref{app:svr_prompt_templates}.

At adaptive inference, the answer returned at the stopping turn $\hat{t}_{\gamma}$ is evaluated using the same task-specific procedure:
\begin{equation}
\widehat{y}_{\gamma}=y_{\hat{t}_{\gamma}}.
\label{eq:app_returned_answer_correctness}
\end{equation}
If no confidence-qualified stopping event occurs, the answer generated at turn $T_{\max}$ is evaluated. Under fixed-budget inference with budget $K$, correctness is computed from the answer produced at turn $K$. No majority voting, best-of-turn selection, reference-assisted choice, or retrospective oracle selection is used in either setting.

\subsection{Metric Definitions}
\label{app:metric_definitions}

Let $\mathcal{E}_d=\{1,\ldots,N_d\}$ denote the evaluation examples of dataset $d$. For example $i$ at turn $t$, let $y_{i,t}\in\{0,1\}$ be the binary task-correctness label defined in Section~\ref{app:answer_evaluation}, $v_{i,t}\in\{\mathrm{C},\mathrm{I},\mathrm{U}\}$ the parsed self-verification verdict, $c_{i,t}\in[0,1]$ the associated confidence, and $z_{i,t}\in\{0,1\}$ an indicator that the response reached the generation-length limit. Under adaptive inference, the returned turn is
\begin{equation}
\begin{aligned}
\hat{t}_i
=
\min\Big(
&\{t\leq T_{\max}:
v_{i,t}=\mathrm{C},\
c_{i,t}\geq\gamma,\
z_{i,t}=0\}\\
&\cup\{T_{\max}\}
\Big).
\end{aligned}
\label{eq:app_returned_turn}
\end{equation}
The union with $\{T_{\max}\}$ ensures that the final allowed turn is returned when no eligible stopping event occurs. A truncated response cannot activate the confidence gate. Under fixed-budget inference with budget $K$, we instead set $\hat{t}_i=K$ for every example.

Final-answer accuracy measures the correctness of the answer returned by the evaluated inference policy:
\begin{equation}
\operatorname{Acc}_d
=
\frac{1}{N_d}
\sum_{i=1}^{N_d}
y_{i,\hat{t}_i}.
\label{eq:app_final_accuracy}
\end{equation}
For adaptive SVR, this is the answer produced at the confidence-gated stopping turn or at $T_{\max}$ when no early stop occurs. For fixed-budget inference, it is the answer produced at the prescribed turn $K$. We do not use majority voting, best-of-turn selection, or retrospective oracle selection.

First-turn accuracy measures performance before refinement:
\begin{equation}
\operatorname{T1Acc}_d
=
\frac{1}{N_d}
\sum_{i=1}^{N_d}
y_{i,1}.
\label{eq:app_first_turn_accuracy}
\end{equation}

Any-turn accuracy measures whether at least one generated answer along the observed trajectory is correct:
\begin{equation}
\operatorname{AnyAcc}_d
=
\frac{1}{N_d}
\sum_{i=1}^{N_d}
\Ind\left[
\max_{1\leq t\leq\hat{t}_i}
y_{i,t}=1
\right].
\label{eq:app_any_turn_accuracy}
\end{equation}
This quantity describes the correction potential of the generated trajectory rather than a deployable answer-selection rule, because identifying the correct turn retrospectively would require oracle correctness labels. Under adaptive inference, turns after $\hat{t}_i$ are not generated and therefore do not contribute to this metric.

We measure inference computation using both the number of generated turns and tokenizer-level sequence cost. Let $p_{i,t}$ and $g_{i,t}$ denote the numbers of prompt and completion tokens, respectively, consumed by example $i$ at turn $t$. Token counts are measured using the backbone tokenizer on the actual model input and generated response at every executed turn. Under adaptive inference, costs are accumulated through $\hat{t}_i$; under fixed-budget inference, they are accumulated over all $K$ turns.

The mean number of generated turns is
\begin{equation}
\operatorname{Turns}_d
=
\frac{1}{N_d}
\sum_{i=1}^{N_d}
\hat{t}_i.
\label{eq:app_mean_turns}
\end{equation}
For a fixed-budget method, $\operatorname{Turns}_d=K$. For adaptive SVR, it reflects the instance-dependent number of refinement steps selected by the stopping controller.

Average prompt-token cost is defined as
\begin{equation}
\operatorname{PromptTok}_d
=
\frac{1}{N_d}
\sum_{i=1}^{N_d}
\sum_{t=1}^{\hat{t}_i}
p_{i,t},
\label{eq:app_avg_prompt_tokens}
\end{equation}
and average completion-token cost is
\begin{equation}
\operatorname{CompletionTok}_d
=
\frac{1}{N_d}
\sum_{i=1}^{N_d}
\sum_{t=1}^{\hat{t}_i}
g_{i,t}.
\label{eq:app_avg_completion_tokens}
\end{equation}
The average total-token cost is
\begin{equation}
\begin{aligned}
\operatorname{TotalTok}_d
&=
\frac{1}{N_d}
\sum_{i=1}^{N_d}
\sum_{t=1}^{\hat{t}_i}
\left(p_{i,t}+g_{i,t}\right)\\
&=
\operatorname{PromptTok}_d
+
\operatorname{CompletionTok}_d.
\end{aligned}
\label{eq:app_avg_total_tokens}
\end{equation}
These quantities correspond to the logged fields \texttt{avg\_prompt\_tokens}, \texttt{avg\_completion\_tokens}, and \texttt{avg\_total\_tokens}. They are cumulative per-example costs over all executed turns rather than per-turn averages. Prompt-token cost includes the complete model input processed at each turn, including the system instruction, original problem, length-bounded previous response, and refinement instruction. Completion-token cost measures newly generated tokens, while total-token cost captures their combined tokenizer-level workload.

Mean turns and token counts characterize complementary aspects of inference cost. Mean turns directly describes the behavior of the adaptive stopping controller, whereas token counts account for differences in prompt and response length that are hidden by turn count alone. In particular, two methods may execute the same number of turns while consuming substantially different numbers of tokens. Token counts are nevertheless workload proxies rather than direct measurements of latency or floating-point operations, since realized runtime also depends on batching, hardware utilization, and inference-engine behavior.

Early Stop Rate measures the fraction of examples that terminate before exhausting the maximum deployment budget:
\begin{equation}
\operatorname{ESR}_d
=
\frac{1}{N_d}
\sum_{i=1}^{N_d}
\Ind[\hat{t}_i<T_{\max}].
\label{eq:app_esr}
\end{equation}
A high ESR indicates that the confidence gate frequently makes an active stopping decision, but does not by itself imply that these decisions are reliable.

Premature Stop Error measures the overall frequency of erroneous early termination:
\begin{equation}
\operatorname{PSE}_d
=
\frac{1}{N_d}
\sum_{i=1}^{N_d}
\Ind[
\hat{t}_i<T_{\max}
\land
y_{i,\hat{t}_i}=0
].
\label{eq:app_pse}
\end{equation}
PSE and ESR use the same denominator $N_d$. PSE is therefore the fraction of all evaluation examples that stop early and return an incorrect answer rather than the error rate conditional on early stopping. When $\operatorname{ESR}_d>0$, the conditional error rate among early-stopped examples is $\operatorname{PSE}_d/\operatorname{ESR}_d$, but this conditional quantity is not used as the primary stopping-risk metric. ESR and PSE are defined for adaptive multi-turn inference and are not directly applicable to fixed-budget methods.

Self-verification diagnostics are computed over all turns generated under the evaluated inference policy. We define the observed-turn set as
\begin{equation}
\mathcal{O}_d
=
\left\{
(i,t):
i\in\mathcal{E}_d,\
1\leq t\leq\hat{t}_i
\right\}.
\label{eq:app_observed_turns}
\end{equation}
Because $\mathcal{O}_d$ depends on the stopping policy, these diagnostics characterize the trajectory distribution induced by the evaluated controller and should be interpreted jointly with final accuracy, mean turns, and token cost.

Verdict accuracy evaluates the discrete correctness judgment made by the model. We interpret \textsc{Correct} as a positive prediction and \textsc{Incorrect} as a negative prediction. Because \textsc{Unsure} represents abstention rather than a binary commitment, it is excluded. Define
\begin{equation}
\mathcal{C}_d
=
\left\{
(i,t)\in\mathcal{O}_d:
v_{i,t}\in\{\mathrm{C},\mathrm{I}\}
\right\}.
\label{eq:app_committed_verdicts}
\end{equation}
Verdict accuracy is
\begin{equation}
\operatorname{VAcc}_d
=
\frac{1}{|\mathcal{C}_d|}
\sum_{(i,t)\in\mathcal{C}_d}
\Ind\left[
\Ind[v_{i,t}=\mathrm{C}]
=
y_{i,t}
\right].
\label{eq:app_verdict_accuracy}
\end{equation}
This metric measures whether committed verdicts distinguish correct from incorrect answers independently of whether confidence reaches the deployment threshold. It is undefined when $\mathcal{C}_d$ is empty.

Probabilistic calibration is measured using the Brier score over all observed turns:
\begin{equation}
\operatorname{Brier}_d
=
\frac{1}{|\mathcal{O}_d|}
\sum_{(i,t)\in\mathcal{O}_d}
(c_{i,t}-y_{i,t})^2.
\label{eq:app_brier}
\end{equation}
Lower values indicate closer agreement between numerical confidence and binary correctness. Unlike final-answer accuracy, the Brier score penalizes both highly confident errors and underconfident correct predictions.

AUROC is computed over $\mathcal{O}_d$ using $c_{i,t}$ as the prediction score and $y_{i,t}$ as the binary label. It measures whether correct answers tend to receive higher confidence than incorrect answers. A value of $0.5$ corresponds to random ranking and a value of $1.0$ to perfect separation. AUROC evaluates ranking quality rather than absolute probability calibration and is undefined when all observed turns have the same correctness label.

Overconfidence measures how frequently the model commits to a \textsc{Correct} verdict despite producing an incorrect answer:
\begin{equation}
\operatorname{Overconf}_d
=
\frac{1}{|\mathcal{O}_d|}
\sum_{(i,t)\in\mathcal{O}_d}
\Ind[
v_{i,t}=\mathrm{C}
\land
y_{i,t}=0
].
\label{eq:app_overconfidence}
\end{equation}
The denominator contains all observed turns rather than only turns predicted \textsc{Correct}, and the metric does not apply the deployment threshold $\gamma$. The training-time asymmetric overconfidence penalty is more fine-grained because it additionally weights each incorrect \textsc{Correct} commitment by its confidence. Overconfidence and PSE capture related but distinct failure modes: Overconf. measures incorrect committed verdicts throughout the observed trajectory, whereas PSE counts only incorrect decisions that actually terminate refinement before $T_{\max}$.

Dataset-level metrics are aggregated through an unweighted macro-average so that every benchmark contributes equally regardless of its sample count. For a dataset collection $\mathcal{D}$ and dataset-level metric $m_d$, we report
\begin{equation}
\operatorname{Macro}(m;\mathcal{D})
=
\frac{1}{|\mathcal{D}|}
\sum_{d\in\mathcal{D}}
m_d.
\label{eq:app_macro_average}
\end{equation}
All-7 contains Countdown, GSM8K, MATH500, AIME26, AMC23, OlympiadBench, and MinervaMath. Math-5 contains MATH500, AIME26, AMC23, OlympiadBench, and MinervaMath. Unless explicitly stated otherwise, accuracy, first-turn accuracy, any-turn accuracy, mean turns, average prompt tokens, average completion tokens, average total tokens, ESR, PSE, verdict accuracy, AUROC, Brier score, and overconfidence are first computed separately for each dataset and then macro-averaged. ESR, PSE, and Overconf. are computed as proportions and displayed as percentages in the result tables.

\section{Additional Experimental Results}

\subsection{Per-Dataset Results}
\label{app:adaptive_per_dataset}

Complete per-dataset behavior under adaptive inference appears in Table~\ref{tab:adaptive_per_dataset}, using the global stopping threshold $\gamma=0.85$, maximum budget $T_{\max}=10$, and greedy decoding. Final and Any denote final-answer and any-turn accuracy, respectively. Turns is the average number of generated turns, while Total Tok. is the average cumulative number of prompt and completion tokens consumed per example over all actually executed turns. ESR, PSE, and Overconf. are reported as percentages. All-7 and Math-5 are unweighted macro-averages of the corresponding dataset-level metrics rather than statistics pooled across examples. Metric definitions follow Section~\ref{app:metric_definitions}.

\begin{table*}[t]
\centering
\setlength{\tabcolsep}{2.4pt}
\caption{Per-dataset results of adaptive SVR with $\gamma=0.85$ and $T_{\max}=10$. Total Tok. denotes the average cumulative token count per example over all executed turns and is reported in thousands. ESR, PSE, and Overconf. are percentages.}
\label{tab:adaptive_per_dataset}
\begin{tabularx}{\textwidth}{@{}>{\raggedright\arraybackslash}p{0.12\textwidth}*{10}{>{\centering\arraybackslash}X}@{}}
\toprule
& \multicolumn{2}{c}{\textbf{Task Quality}}
& \multicolumn{2}{c}{\textbf{Inference Cost}}
& \multicolumn{2}{c}{\textbf{Stopping Behavior}}
& \multicolumn{4}{c}{\textbf{Self-Verification Quality}} \\
\cmidrule(lr){2-3}
\cmidrule(lr){4-5}
\cmidrule(lr){6-7}
\cmidrule(lr){8-11}
\textbf{Dataset}
& \textbf{Final}
& \textbf{Any}
& \textbf{Turns}
& \textbf{\shortstack{Total Tok.\\($\times10^3$)}}
& \textbf{ESR}
& \textbf{PSE}
& \textbf{V-Acc}
& \textbf{AUROC}
& \textbf{Brier}
& \textbf{Overconf.} \\
\midrule
Countdown       & 0.839 & 0.849 & 2.69 & 3.46  & 89.7\%  & 5.8\%  & 0.935 & 0.822 & 0.095 & 2.3\%  \\
GSM8K           & 0.813 & 0.820 & 1.12 & 1.02  & 100.0\% & 18.7\% & 0.813 & 0.505 & 0.184 & 17.6\% \\
MATH500         & 0.676 & 0.684 & 2.33 & 4.83  & 90.8\%  & 23.2\% & 0.744 & 0.651 & 0.211 & 15.6\% \\
AIME26          & 0.200 & 0.200 & 6.40 & 23.84 & 60.0\%  & 40.0\% & 0.333 & 0.750 & 0.342 & 15.3\% \\
AMC23           & 0.550 & 0.550 & 3.08 & 7.14  & 85.0\%  & 30.0\% & 0.647 & 0.667 & 0.245 & 14.8\% \\
OlympiadBench   & 0.417 & 0.430 & 3.74 & 16.18 & 80.1\%  & 38.4\% & 0.518 & 0.672 & 0.336 & 23.6\% \\
MinervaMath     & 0.449 & 0.463 & 1.56 & 3.47  & 98.2\%  & 53.3\% & 0.457 & 0.543 & 0.482 & 46.1\% \\
\midrule
All-7           & 0.563 & 0.571 & 2.99 & 8.56  & 86.3\%  & 29.9\% & 0.635 & 0.659 & 0.271 & 19.3\% \\
Math-5          & 0.458 & 0.465 & 3.42 & 11.09 & 82.8\%  & 37.0\% & 0.540 & 0.657 & 0.323 & 23.1\% \\
\bottomrule
\end{tabularx}
\end{table*}

The results reveal substantial variation in adaptive computation across benchmarks. Countdown achieves the highest final accuracy of $0.839$ while requiring only $2.69$ turns and $3.46$ thousand tokens per example. GSM8K incurs the lowest inference cost, using $1.12$ turns and $1.02$ thousand tokens, consistent with the limited refinement required by this comparatively saturated benchmark. In contrast, AIME26 consumes $6.40$ turns and $23.84$ thousand tokens per example, showing that the controller allocates substantially more computation to difficult problems. OlympiadBench also incurs a high token cost of $16.18$ thousand despite using only $3.74$ turns, demonstrating that turn count alone does not capture differences in reasoning and prompt length.

Across all seven datasets, adaptive SVR uses $2.99$ turns and $8.56$ thousand tokens per example on average. The corresponding Math-5 costs increase to $3.42$ turns and $11.09$ thousand tokens, reflecting the greater reasoning demands of the mathematical benchmarks. Mean turns characterize the stopping decisions of the adaptive controller, whereas cumulative token cost additionally captures variation in problem length, refinement-prompt length, and generated reasoning length.

The stopping diagnostics show that inexpensive inference need not imply low stopping error. MinervaMath terminates after only $1.56$ turns and $3.47$ thousand tokens on average, but its PSE reaches $53.3\%$, its Brier score is $0.482$, and its overconfidence rate is $46.1\%$. Countdown presents the opposite observed pattern: ESR is $89.7\%$, PSE is $5.8\%$, Brier score is $0.095$, and overconfidence is $2.3\%$. Thus, stopping quality varies substantially across benchmarks.

The gap between any-turn and final-answer accuracy remains small across all datasets, ranging from zero to $1.4$ percentage points. This indicates that adaptive SVR retains most correct answers encountered along its generated trajectories without majority voting, best-of-turn selection, or retrospective access to oracle correctness. Nevertheless, the substantial variation in PSE, calibration, and token consumption confirms that adaptive inference should be evaluated jointly in terms of answer quality, stopping error, and cumulative computational cost.

\subsection{Fixed-Budget and Threshold Sweeps}
\label{app:budget_threshold_sweeps}

\noindent\textbf{Fixed-budget turn sweep.}\quad\label{app:fixed_budget_sweep} To isolate the effect of assigning a uniform refinement budget, we disable confidence-gated stopping and require every example to execute exactly $K\in\{1,\ldots,10\}$ turns. The answer produced at turn $K$ is returned as the final prediction. All generations use greedy decoding. We generate one complete ten-turn trajectory for each example and evaluate every prefix by selecting its final turn; under deterministic decoding, later generations cannot alter earlier outputs, so this procedure is equivalent to independently executing each fixed budget. No confidence gate, majority voting, best-of-turn selection, or retrospective oracle selection is used. This experiment isolates the effect of a shared stopping position, while the tokenizer-level cost of the adaptive operating point is reported separately in Section~\ref{app:adaptive_per_dataset}.

\begin{table*}[t]
\centering
\setlength{\tabcolsep}{2.6pt}
\caption{Final-answer accuracy of SVR under fixed-budget inference. Every example is forced to execute exactly $K$ turns, and the answer generated at turn $K$ is returned. All-7 and Math-5 are unweighted macro-averages defined in Section~\ref{app:metric_definitions}. Bold values identify the best fixed-budget result in each column. The adaptive row, shown only for reference, uses $\gamma=0.85$ and $T_{\max}=10$ and is not included when identifying the fixed-budget optima.}
\label{tab:fixed_budget_sweep}
\begin{tabular*}{\textwidth}{@{\extracolsep{\fill}}lccccccccc@{}}
\toprule
& \multicolumn{7}{c}{\textbf{Per-Dataset Final Accuracy}}
& \multicolumn{2}{c}{\textbf{Macro Average}} \\
\cmidrule(lr){2-8}
\cmidrule(lr){9-10}
\textbf{Budget}
& \textbf{Countdown}
& \textbf{GSM8K}
& \textbf{MATH500}
& \textbf{AIME26}
& \textbf{AMC23}
& \textbf{\shortstack{Oly.}}
& \textbf{\shortstack{Minerva}}
& \textbf{All-7}
& \textbf{Math-5} \\
\midrule
$K=1$  & 0.667 & \textbf{0.798} & \textbf{0.578} & 0.033 & 0.350 & \textbf{0.319} & \textbf{0.401} & 0.449 & \textbf{0.336} \\
$K=2$  & 0.692 & 0.682 & 0.452 & 0.067 & 0.375 & 0.258 & 0.309 & 0.405 & 0.292 \\
$K=3$  & 0.700 & 0.719 & 0.494 & 0.067 & 0.375 & 0.252 & 0.298 & 0.415 & 0.297 \\
$K=4$  & 0.720 & 0.705 & 0.524 & 0.067 & 0.350 & 0.276 & 0.283 & 0.418 & 0.300 \\
$K=5$  & 0.706 & 0.732 & 0.528 & 0.067 & 0.375 & 0.270 & 0.290 & 0.424 & 0.306 \\
$K=6$  & 0.718 & 0.721 & 0.518 & 0.033 & 0.350 & 0.269 & 0.298 & 0.415 & 0.294 \\
$K=7$  & 0.734 & 0.726 & 0.554 & 0.033 & 0.400 & 0.273 & 0.294 & 0.431 & 0.311 \\
$K=8$  & 0.741 & 0.723 & 0.512 & 0.067 & \textbf{0.450} & 0.277 & 0.331 & 0.443 & 0.327 \\
$K=9$  & \textbf{0.749} & 0.741 & 0.538 & 0.067 & 0.400 & 0.263 & 0.316 & 0.439 & 0.317 \\
$K=10$ & 0.742 & 0.736 & 0.570 & \textbf{0.133} & 0.375 & 0.292 & 0.305 & \textbf{0.450} & 0.335 \\
\midrule
\textbf{Adaptive}
& 0.839 & 0.813 & 0.676 & 0.200 & 0.550 & 0.417 & 0.449 & 0.563 & 0.458 \\
\bottomrule
\end{tabular*}
\end{table*}

The fixed-budget sweep in Table~\ref{tab:fixed_budget_sweep} is strongly non-monotonic. All-7 accuracy decreases from $0.449$ at $K=1$ to $0.405$ at $K=2$ and reaches its highest fixed-budget value of $0.450$ only at $K=10$. Math-5 follows a different pattern: its best fixed-budget accuracy, $0.336$, occurs at $K=1$, and none of the larger shared budgets produces a consistent improvement. The optimal stopping position also varies substantially across individual benchmarks. Countdown benefits from extended refinement and peaks at $K=9$, AIME26 peaks at $K=10$, and AMC23 peaks at $K=8$. In contrast, GSM8K, MATH500, OlympiadBench, and MinervaMath obtain their highest fixed-budget accuracies at the first turn. These heterogeneous optima indicate that no single turn budget is uniformly appropriate across tasks or examples.

Forced continuation can overwrite previously correct answers. The clearest deterioration occurs between the first and second turns: GSM8K decreases from $0.798$ to $0.682$, a loss of $11.6$ percentage points, while MATH500 decreases from $0.578$ to $0.452$, a loss of $12.6$ points. This behavior does not imply that subsequent turns are incapable of solving additional examples. On GSM8K, for example, any-turn accuracy increases from $0.798$ at $K=1$ to $0.925$ over the complete ten-turn trajectory, whereas the answer returned specifically at turn ten achieves only $0.736$. The model therefore encounters correct solutions for additional examples during refinement but cannot reliably preserve them at a single globally prescribed turn.

Adaptive SVR avoids committing to a shared stopping position and instead returns an instance-dependent answer using its own self-verification signal. It achieves an All-7 accuracy of $0.563$, exceeding the best fixed-budget result of $0.450$ by $11.3$ percentage points, while using only $2.99$ turns and $8.56$ thousand tokens per example on average. On Math-5, adaptive SVR reaches $0.458$, exceeding the best fixed-budget accuracy of $0.336$ by $12.2$ points, with an average cost of $3.42$ turns and $11.09$ thousand tokens. Adaptive inference also exceeds the best post-hoc fixed-turn result separately on every benchmark, with gains ranging from $1.5$ points on GSM8K to $10.0$ points on AMC23.

These results distinguish adaptive answer retention from simply increasing the refinement budget. A larger fixed budget can expose additional correct intermediate solutions, but it also forces already solved examples to undergo further revisions and provides no mechanism for selecting the appropriate stopping position for each trajectory. SVR instead uses self-verification to preserve confident solutions while reserving additional turns for unresolved examples. The resulting advantage therefore arises from instance-dependent compute allocation rather than from uniformly extending every reasoning trajectory.

\noindent\textbf{Confidence-threshold sweep.}\quad\label{app:confidence_threshold_sweep} We evaluate the sensitivity of adaptive SVR to the confidence threshold by sweeping $\gamma\in\{0.50,0.55,\\\ldots,0.95\}$ while fixing the maximum inference budget at $T_{\max}=10$. Every configuration uses greedy decoding and applies the same global threshold across all seven benchmarks. SVR stops at the first non-truncated turn satisfying $v_t=\mathrm{C}$ and $c_t\geq\gamma$; otherwise, it returns the answer produced at turn $T_{\max}$. The shared threshold $\gamma=0.85$ is used for all main results and is not selected separately for individual benchmarks.

\begin{table*}[t]
\centering
\setlength{\tabcolsep}{2.8pt}
\caption{Final-answer accuracy of adaptive SVR under different confidence thresholds. All configurations use $T_{\max}=10$ and greedy decoding. All-7 and Math-5 are unweighted macro-averages defined in Section~\ref{app:metric_definitions}.}
\label{tab:confidence_threshold_accuracy}
\begin{tabular*}{\textwidth}{@{\extracolsep{\fill}}lccccccccc@{}}
\toprule
\textbf{$\gamma$}
& \textbf{Countdown}
& \textbf{GSM8K}
& \textbf{MATH500}
& \textbf{AIME26}
& \textbf{AMC23}
& \textbf{\shortstack{Oly.}}
& \textbf{\shortstack{Minerva}}
& \textbf{All-7}
& \textbf{Math-5} \\
\midrule
0.50 & 0.831 & 0.810 & 0.672 & 0.100 & 0.550 & 0.402 & 0.430 & 0.542 & 0.431 \\
0.55 & 0.829 & 0.812 & 0.652 & 0.100 & 0.550 & 0.377 & 0.412 & 0.533 & 0.418 \\
0.60 & 0.836 & 0.809 & 0.666 & 0.100 & 0.525 & 0.396 & 0.438 & 0.539 & 0.425 \\
0.65 & 0.842 & 0.809 & 0.670 & 0.100 & 0.550 & 0.374 & 0.412 & 0.537 & 0.421 \\
0.70 & 0.832 & 0.806 & 0.674 & 0.133 & 0.475 & 0.402 & 0.401 & 0.532 & 0.417 \\
0.75 & 0.835 & 0.804 & 0.668 & 0.100 & 0.500 & 0.392 & 0.423 & 0.532 & 0.417 \\
0.80 & 0.835 & 0.808 & 0.666 & 0.133 & 0.525 & 0.380 & 0.423 & 0.539 & 0.425 \\
0.85 & 0.839 & 0.813 & 0.676 & 0.200 & 0.550 & 0.417 & 0.449 & 0.563 & 0.458 \\
0.90 & 0.822 & 0.811 & 0.660 & 0.033 & 0.500 & 0.401 & 0.404 & 0.519 & 0.400 \\
0.95 & 0.829 & 0.803 & 0.664 & 0.100 & 0.525 & 0.389 & 0.426 & 0.534 & 0.421 \\
\bottomrule
\end{tabular*}
\end{table*}

Within the threshold grid reported in Table~\ref{tab:confidence_threshold_accuracy}, the shared setting $\gamma=0.85$ attains the highest aggregate accuracy, reaching $0.563$ on All-7 and $0.458$ on Math-5. The same global setting obtains the highest observed accuracy on GSM8K, MATH500, AIME26, OlympiadBench, and MinervaMath, and ties the highest result on AMC23. Countdown reaches its maximum at $\gamma=0.65$, but the improvement over the default is only $0.3$ percentage points. The selected threshold therefore transfers reasonably across benchmarks without dataset-specific tuning.

\begin{table*}[t]
\centering
\setlength{\tabcolsep}{2.2pt}
\caption{Macro-averaged operating characteristics of adaptive SVR across confidence thresholds. Turns denotes mean inference turns, and Total Tok. denotes the average cumulative prompt and completion token count per example over all executed turns, reported in thousands. ESR is Early Stop Rate, and PSE is Premature Stop Error; both are reported as percentages.}
\label{tab:confidence_threshold_operating}
\begin{tabular*}{\textwidth}{@{\extracolsep{\fill}}lcccccccccc@{}}
\toprule
& \multicolumn{2}{c}{\textbf{Accuracy}}
& \multicolumn{2}{c}{\textbf{Turns}}
& \multicolumn{2}{c}{\textbf{Total Tok. ($\times10^3$)}}
& \multicolumn{2}{c}{\textbf{ESR}}
& \multicolumn{2}{c}{\textbf{PSE}} \\
\cmidrule(lr){2-3}
\cmidrule(lr){4-5}
\cmidrule(lr){6-7}
\cmidrule(lr){8-9}
\cmidrule(lr){10-11}
\textbf{$\gamma$}
& \textbf{All-7}
& \textbf{Math-5}
& \textbf{All-7}
& \textbf{Math-5}
& \textbf{All-7}
& \textbf{Math-5}
& \textbf{All-7}
& \textbf{Math-5}
& \textbf{All-7}
& \textbf{Math-5} \\
\midrule
0.50 & 0.542 & 0.431 & 2.99 & 3.46 & 8.54 & 11.10 & 84.8\% & 80.7\% & 30.6\% & 37.7\% \\
0.55 & 0.533 & 0.418 & 3.00 & 3.44 & 8.51 & 11.03 & 85.3\% & 81.4\% & 32.0\% & 39.6\% \\
0.60 & 0.539 & 0.425 & 2.94 & 3.36 & 8.30 & 10.75 & 86.1\% & 82.4\% & 32.3\% & 40.0\% \\
0.65 & 0.537 & 0.421 & 2.85 & 3.26 & 8.10 & 10.48 & 86.7\% & 83.1\% & 33.1\% & 41.1\% \\
0.70 & 0.532 & 0.417 & 2.97 & 3.40 & 8.49 & 11.00 & 85.8\% & 82.1\% & 33.2\% & 41.1\% \\
0.75 & 0.532 & 0.417 & 2.95 & 3.37 & 8.40 & 10.85 & 85.7\% & 82.0\% & 32.5\% & 40.4\% \\
0.80 & 0.539 & 0.425 & 2.91 & 3.32 & 8.21 & 10.61 & 85.0\% & 81.1\% & 31.2\% & 38.5\% \\
0.85
& 0.563
& 0.458
& 2.99
& 3.42
& 8.56
& 11.09
& 86.3\%
& 82.8\%
& 29.9\%
& 37.0\% \\
0.90 & 0.519 & 0.400 & 3.15 & 3.63 & 9.13 & 11.88 & 83.1\% & 78.3\% & 31.2\% & 38.4\% \\
0.95 & 0.534 & 0.421 & 2.97 & 3.39 & 8.48 & 10.96 & 84.3\% & 80.2\% & 31.0\% & 38.1\% \\
\bottomrule
\end{tabular*}
\end{table*}

The operating-point comparison in Table~\ref{tab:confidence_threshold_operating} demonstrates that the minimum-compute threshold is not the strongest choice for task performance or stopping error. At $\gamma=0.65$, SVR uses the lowest observed costs of $8.10$ thousand tokens on All-7 and $10.48$ thousand tokens on Math-5, but its aggregate accuracies are only $0.537$ and $0.421$, while PSE rises to $33.1\%$ and $41.1\%$. In comparison, the default $\gamma=0.85$ uses $8.56$ and $11.09$ thousand tokens while improving accuracy to $0.563$ and $0.458$ and reducing PSE to $29.9\%$ and $37.0\%$. Relative to the minimum-token setting, the default requires only $0.46$ thousand additional tokens on All-7 and $0.61$ thousand on Math-5, while gaining $2.6$ and $3.7$ percentage points in accuracy.

A stricter confidence threshold does not necessarily reduce erroneous stopping or improve task accuracy. Raising the threshold from $\gamma=0.85$ to $\gamma=0.90$ increases mean turns from $2.99$ to $3.15$ on All-7 and from $3.42$ to $3.63$ on Math-5. Total-token consumption correspondingly increases from $8.56$ to $9.13$ thousand and from $11.09$ to $11.88$ thousand. Despite this additional computation, All-7 accuracy decreases from $0.563$ to $0.519$, Math-5 accuracy decreases from $0.458$ to $0.400$, and PSE increases on both aggregates. Delaying termination can therefore expose a correct intermediate solution to additional refinement that fails to preserve its correctness.

The relationship between the confidence threshold and computation is non-monotonic. Although increasing $\gamma$ makes the stopping criterion more difficult to satisfy for a fixed trajectory, the trajectories observed at different thresholds can vary in both their executed turns and their prompt and completion lengths. Consequently, configurations with similar mean turns may incur different cumulative token costs, and neither Turns nor Total Tok. changes monotonically across the sweep. For example, $\gamma=0.50$ and $\gamma=0.85$ both use $2.99$ All-7 turns on average, but consume $8.54$ and $8.56$ thousand tokens, respectively.

Threshold sensitivity also varies across benchmarks. Countdown, GSM8K, and MATH500 remain comparatively stable over the evaluated grid, with accuracy ranges of $2.0$, $1.0$, and $2.4$ percentage points, respectively. AMC23, OlympiadBench, and MinervaMath exhibit larger ranges of $7.5$, $4.3$, and $4.8$ points. AIME26 is the most sensitive benchmark, ranging from $0.033$ to $0.200$ accuracy, although this variation should be interpreted cautiously because the dataset contains only $30$ problems. In particular, its accuracy decreases from $0.200$ at $\gamma=0.85$ to $0.033$ at $\gamma=0.90$, indicating that an excessively conservative threshold may defer termination until later, less reliable turns on difficult examples.

Overall, the sweep is not a monotonic trade-off in which a larger threshold always reduces stopping error at the cost of more computation. Lower thresholds can permit incorrect confident termination, whereas an excessively high threshold may consume more compute while failing to preserve correct intermediate solutions. The shared setting $\gamma=0.85$ is the strongest observed aggregate operating point in this diagnostic grid. All main experiments use this one global threshold, without substituting benchmark-specific optima.

\subsection{Complete Ablation Results}
\label{app:complete_ablations}

\noindent\textbf{Reward-component ablations.}\quad We report the complete per-dataset reward-component results in Tables~\ref{tab:app_reward_acc} and~\ref{tab:app_reward_diag}. All variants use the same backbone, domain-specific training data, optimization configuration, structured self-check interface, and greedy decoding protocol. Adaptive inference uses the shared threshold $\gamma=0.85$ and maximum budget $T_{\max}=10$. The variant without $R_{\mathrm{verify}}$ retains the trajectory-averaged solve and format rewards but removes the complete self-verification block, comprising calibration, asymmetric overconfidence control, error detection, and stop readiness. The remaining variants independently remove $R_{\mathrm{cal}}$, $R_{\mathrm{over}}$, or $R_{\mathrm{ready}}$. We do not separately ablate $R_{\mathrm{detect}}$; its contribution is removed only as part of the complete $R_{\mathrm{verify}}$ ablation. Because cumulative token counts were not retained for every ablation run, compute comparisons in this subsection use mean inference turns rather than extrapolated token costs.

\begin{table*}[t]
\centering
\renewcommand{\arraystretch}{1.06}
\caption{Per-dataset final-answer accuracy of the reward-component ablations. All variants use adaptive inference with $\gamma=0.85$ and $T_{\max}=10$. All-7 and Math-5 are unweighted macro-averages.}
\label{tab:app_reward_acc}
\begin{tabular*}{\textwidth}{@{\extracolsep{\fill}}lccccccccc@{}}
\toprule
Variant
& Countdown
& GSM8K
& MATH500
& AIME26
& AMC23
& Oly.
& Minerva
& All-7
& Math-5 \\
\midrule
Full SVR
& 0.839
& 0.813
& 0.676
& 0.200
& 0.550
& 0.417
& 0.449
& 0.563
& 0.458 \\
w/o $R_{\mathrm{verify}}$
& 0.829
& 0.794
& 0.670
& 0.100
& 0.500
& 0.395
& 0.408
& 0.528
& 0.415 \\
w/o $R_{\mathrm{cal}}$
& 0.790
& 0.801
& 0.672
& 0.133
& 0.525
& 0.387
& 0.441
& 0.536
& 0.432 \\
w/o $R_{\mathrm{over}}$
& 0.722
& 0.778
& 0.682
& 0.133
& 0.475
& 0.401
& 0.434
& 0.518
& 0.425 \\
w/o $R_{\mathrm{ready}}$
& 0.758
& 0.486
& 0.654
& 0.167
& 0.575
& 0.398
& 0.423
& 0.494
& 0.443 \\
\bottomrule
\end{tabular*}
\end{table*}

Table~\ref{tab:app_reward_acc} shows that Full SVR achieves the highest aggregate accuracy, reaching $0.563$ on All-7 and $0.458$ on Math-5, and obtains the strongest result on five of the seven individual benchmarks. Removing the complete self-verification block decreases these macro-averages by $3.5$ and $4.3$ percentage points, respectively, indicating that solve-related shaping and the structured output interface alone do not produce an equally effective control signal. Each independently evaluated reward removal also reduces both aggregate accuracy measures.

The only per-dataset exceptions are MATH500, where removing $R_{\mathrm{over}}$ increases accuracy from $0.676$ to $0.682$, and AMC23, where removing $R_{\mathrm{ready}}$ increases accuracy from $0.550$ to $0.575$. The MATH500 difference is $0.6$ percentage points, while the AMC23 difference corresponds to approximately one additional correct answer on its 40-example evaluation set. Neither local improvement transfers to the remaining benchmarks or either macro-average.

\begin{table*}[t]
\centering
\renewcommand{\arraystretch}{1.03}
\caption{Complete stopping and confidence diagnostics for the reward-component ablations. Turns is the mean number of executed inference turns. ESR is the fraction of examples stopped before $T_{\max}$, and PSE is the fraction of all examples that stop before $T_{\max}$ and return an incorrect answer. Brier is the mean squared error between confidence and binary correctness. Overconf. is the fraction of observed turns on which an incorrect answer receives a \textsc{Correct} verdict. ESR, PSE, and Overconf. are percentages. All-7 and Math-5 are unweighted macro-averages.}
\label{tab:app_reward_diag}
\begin{tabular*}{\textwidth}{@{\extracolsep{\fill}}llccccccccc@{}}
\toprule
Metric
& Variant
& Countdown
& GSM8K
& MATH500
& AIME26
& AMC23
& Oly.
& Minerva
& All-7
& Math-5 \\
\midrule
\multirow{5}{*}{Turns}
& Full SVR
& 2.69 & 1.12 & 2.33 & 6.40 & 3.08 & 3.74 & 1.56 & 2.99 & 3.42 \\
& w/o $R_{\mathrm{verify}}$
& 2.75 & 1.05 & 2.23 & 6.93 & 3.78 & 3.68 & 1.52 & 3.13 & 3.63 \\
& w/o $R_{\mathrm{cal}}$
& 2.76 & 1.58 & 2.29 & 6.13 & 3.75 & 3.78 & 1.46 & 3.11 & 3.48 \\
& w/o $R_{\mathrm{over}}$
& 2.36 & 1.09 & 2.43 & 6.67 & 3.70 & 3.85 & 1.40 & 3.07 & 3.61 \\
& w/o $R_{\mathrm{ready}}$
& 3.47 & 5.75 & 2.29 & 6.27 & 3.50 & 3.89 & 1.61 & 3.83 & 3.51 \\
\midrule
\multirow{5}{*}{ESR}
& Full SVR
& 89.7 & 100.0 & 90.8 & 60.0 & 85.0 & 80.1 & 98.2 & 86.3 & 82.8 \\
& w/o $R_{\mathrm{verify}}$
& 85.8 & 99.9 & 91.4 & 46.7 & 77.5 & 80.9 & 98.5 & 83.0 & 79.0 \\
& w/o $R_{\mathrm{cal}}$
& 91.0 & 100.0 & 91.2 & 66.7 & 85.0 & 79.1 & 98.9 & 87.4 & 84.2 \\
& w/o $R_{\mathrm{over}}$
& 97.3 & 100.0 & 89.4 & 50.0 & 85.0 & 77.0 & 100.0 & 85.5 & 80.3 \\
& w/o $R_{\mathrm{ready}}$
& 81.1 & 50.3 & 91.0 & 56.7 & 82.5 & 77.2 & 97.8 & 76.6 & 81.0 \\
\midrule
\multirow{5}{*}{PSE}
& Full SVR
& 5.8 & 18.7 & 23.2 & 40.0 & 30.0 & 38.4 & 53.3 & 29.9 & 37.0 \\
& w/o $R_{\mathrm{verify}}$
& 3.0 & 20.5 & 24.4 & 36.7 & 27.5 & 41.8 & 57.7 & 30.2 & 37.6 \\
& w/o $R_{\mathrm{cal}}$
& 12.0 & 19.9 & 24.0 & 53.3 & 32.5 & 40.4 & 54.8 & 33.8 & 41.0 \\
& w/o $R_{\mathrm{over}}$
& 25.1 & 22.2 & 21.2 & 36.7 & 37.5 & 37.1 & 56.6 & 33.8 & 37.8 \\
& w/o $R_{\mathrm{ready}}$
& 5.3 & 11.7 & 25.6 & 40.0 & 25.0 & 37.4 & 55.5 & 28.6 & 36.7 \\
\midrule
\multirow{5}{*}{Brier}
& Full SVR
& 0.095 & 0.184 & 0.211 & 0.342 & 0.245 & 0.336 & 0.482 & 0.271 & 0.323 \\
& w/o $R_{\mathrm{verify}}$
& 0.077 & 0.201 & 0.225 & 0.372 & 0.270 & 0.361 & 0.512 & 0.288 & 0.348 \\
& w/o $R_{\mathrm{cal}}$
& 0.118 & 0.187 & 0.221 & 0.378 & 0.283 & 0.345 & 0.493 & 0.289 & 0.344 \\
& w/o $R_{\mathrm{over}}$
& 0.179 & 0.214 & 0.199 & 0.383 & 0.283 & 0.341 & 0.509 & 0.301 & 0.343 \\
& w/o $R_{\mathrm{ready}}$
& 0.116 & 0.272 & 0.229 & 0.348 & 0.258 & 0.335 & 0.496 & 0.293 & 0.333 \\
\midrule
\multirow{5}{*}{Overconf.}
& Full SVR
& 2.3 & 17.6 & 15.6 & 15.3 & 14.8 & 23.6 & 46.1 & 19.3 & 23.1 \\
& w/o $R_{\mathrm{verify}}$
& 1.3 & 19.8 & 17.1 & 18.6 & 17.2 & 26.1 & 49.3 & 21.4 & 25.7 \\
& w/o $R_{\mathrm{cal}}$
& 4.6 & 17.9 & 16.9 & 19.5 & 17.5 & 24.6 & 47.5 & 21.2 & 25.2 \\
& w/o $R_{\mathrm{over}}$
& 11.3 & 21.0 & 14.1 & 20.8 & 17.4 & 24.0 & 49.0 & 22.5 & 25.1 \\
& w/o $R_{\mathrm{ready}}$
& 2.5 & 18.7 & 17.5 & 16.8 & 15.5 & 23.0 & 47.5 & 20.2 & 24.1 \\
\bottomrule
\end{tabular*}
\end{table*}

The stop-readiness ablation produces the clearest failure to convert solved states into actionable stopping decisions. Removing $R_{\mathrm{ready}}$ increases All-7 computation from $2.99$ to $3.83$ turns and reduces ESR from $86.3\%$ to $76.6\%$, while All-7 accuracy decreases by $6.9$ percentage points. The effect is concentrated on GSM8K, where mean computation increases from $1.12$ to $5.75$ turns, ESR falls from $100.0\%$ to $50.3\%$, and final accuracy decreases from $0.813$ to $0.486$. Its lower GSM8K PSE does not indicate safer control: the model incurs fewer premature-stop errors primarily because it makes substantially fewer early-stop decisions, while the additional refinement fails to preserve or recover task performance.

Removing $R_{\mathrm{over}}$ produces a different failure mode. On Countdown, mean turns decrease from $2.69$ to $2.36$ and ESR increases from $89.7\%$ to $97.3\%$, but PSE rises from $5.8\%$ to $25.1\%$ and final accuracy falls from $0.839$ to $0.722$. The All-7 overconfidence rate correspondingly increases from $19.3\%$ to $22.5\%$. Without the asymmetric penalty, incorrect answers are more frequently accompanied by positive commitments that can activate the stopping gate.

Removing $R_{\mathrm{cal}}$ causes a broader degradation in confidence quality. The All-7 Brier score increases from $0.271$ to $0.289$, overconfidence rises from $19.3\%$ to $21.2\%$, and PSE rises from $29.9\%$ to $33.8\%$. On Math-5, PSE increases from $37.0\%$ to $41.0\%$. The largest per-dataset increase occurs on AIME26, where PSE rises from $40.0\%$ to $53.3\%$, although this estimate should be interpreted cautiously because the benchmark contains only 30 evaluation examples. Together, the $R_{\mathrm{cal}}$ and $R_{\mathrm{over}}$ ablations support complementary interpretations: calibration broadly aligns numerical confidence with binary correctness, whereas asymmetric overconfidence control targets incorrect commitments that are particularly hazardous for adaptive stopping.

Individual diagnostics must be interpreted jointly with task accuracy and stopping frequency. For example, removing the complete $R_{\mathrm{verify}}$ block lowers the Countdown Brier score from $0.095$ to $0.077$ and its overconfidence rate from $2.3\%$ to $1.3\%$, yet reduces aggregate final accuracy. Such isolated improvements can result from conservative or weakly committed self-assessments rather than from a more useful controller. Across the reward ablations, no variant simultaneously matches Full SVR in aggregate accuracy, compute allocation, calibration quality, and stopping behavior.

\noindent\textbf{Structural and controller-interface ablations.}\quad We next examine whether SVR requires iterative refinement and whether either field of its structured self-check is sufficient in isolation. Single-turn SVR retains the structured self-check and applicable one-turn reward terms but is trained and evaluated with a fixed one-turn horizon. The verdict-only and confidence-only policies are independently retrained and emit only the retained self-verification field. Verdict-only retains discrete error-detection supervision and stops on a non-truncated \textsc{Correct} verdict, whereas confidence-only retains Brier-style calibration and stops when $c_t\geq\gamma$. All other training and inference settings remain unchanged.

\begin{table*}[t]
\centering
\renewcommand{\arraystretch}{1.03}
\caption{Complete structural and controller-interface ablations. Full SVR emits both verdict and confidence. Single-turn SVR executes exactly one turn and therefore has no adaptive multi-turn stopping decision. Verdict-only and confidence-only are independently trained and use their corresponding single-signal stopping rules. ESR and PSE are percentages. All-7 and Math-5 are unweighted macro-averages.}
\label{tab:app_interface}
\begin{tabular*}{\textwidth}{@{\extracolsep{\fill}}llccccccccc@{}}
\toprule
Metric
& Variant
& Countdown
& GSM8K
& MATH500
& AIME26
& AMC23
& Oly.
& Minerva
& All-7
& Math-5 \\
\midrule
\multirow{4}{*}{Acc.}
& Full SVR
& 0.839 & 0.813 & 0.676 & 0.200 & 0.550 & 0.417 & 0.449 & 0.563 & 0.458 \\
& Single-turn SVR
& 0.611 & 0.810 & 0.624 & 0.100 & 0.400 & 0.359 & 0.390 & 0.471 & 0.375 \\
& Verdict-only
& 0.771 & 0.741 & 0.688 & 0.100 & 0.575 & 0.420 & 0.408 & 0.529 & 0.438 \\
& Confidence-only
& 0.659 & 0.795 & 0.616 & 0.100 & 0.475 & 0.356 & 0.375 & 0.482 & 0.384 \\
\midrule
\multirow{4}{*}{Turns}
& Full SVR
& 2.69 & 1.12 & 2.33 & 6.40 & 3.08 & 3.74 & 1.56 & 2.99 & 3.42 \\
& Single-turn SVR
& 1.00 & 1.00 & 1.00 & 1.00 & 1.00 & 1.00 & 1.00 & 1.00 & 1.00 \\
& Verdict-only
& 3.04 & 1.15 & 2.18 & 5.93 & 3.58 & 3.73 & 1.35 & 2.99 & 3.35 \\
& Confidence-only
& 3.21 & 6.53 & 5.02 & 8.47 & 7.15 & 7.07 & 4.46 & 5.99 & 6.43 \\
\midrule
\multirow{4}{*}{ESR}
& Full SVR
& 89.7 & 100.0 & 90.8 & 60.0 & 85.0 & 80.1 & 98.2 & 86.3 & 82.8 \\
& Single-turn SVR
& -- & -- & -- & -- & -- & -- & -- & -- & -- \\
& Verdict-only
& 81.1 & 100.0 & 91.8 & 63.3 & 82.5 & 78.2 & 99.3 & 85.2 & 83.0 \\
& Confidence-only
& 83.6 & 61.6 & 67.4 & 26.7 & 45.0 & 43.2 & 76.8 & 57.8 & 51.8 \\
\midrule
\multirow{4}{*}{PSE}
& Full SVR
& 5.8 & 18.7 & 23.2 & 40.0 & 30.0 & 38.4 & 53.3 & 29.9 & 37.0 \\
& Single-turn SVR
& -- & -- & -- & -- & -- & -- & -- & -- & -- \\
& Verdict-only
& 4.0 & 25.9 & 23.0 & 53.3 & 25.0 & 36.4 & 58.5 & 32.3 & 39.2 \\
& Confidence-only
& 17.8 & 13.3 & 17.4 & 20.0 & 10.0 & 19.3 & 45.2 & 20.4 & 22.4 \\
\bottomrule
\end{tabular*}
\end{table*}

Single-turn SVR decreases All-7 accuracy from $0.563$ to $0.471$ and Math-5 accuracy from $0.458$ to $0.375$. The largest loss occurs on Countdown, where accuracy falls by $22.8$ percentage points. GSM8K is the principal exception because its first-turn performance is already close to the adaptive result, leaving comparatively little room for refinement. The aggregate gap nevertheless shows that structured self-assessment alone does not account for SVR's gains; access to repeated correction and answer preservation is itself necessary.

Verdict-only reduces All-7 accuracy to $0.529$ and Math-5 accuracy to $0.438$. Although its aggregate turn count and ESR remain close to those of Full SVR, this similarity conceals task-dependent behavior. On GSM8K, verdict-only terminates after $1.15$ turns on average but decreases accuracy from $0.813$ to $0.741$. On Countdown, it instead increases computation from $2.69$ to $3.04$ turns while accuracy still decreases from $0.839$ to $0.771$. The verdict supplies a categorical commitment, but without confidence the controller lacks a continuous and threshold-adjustable notion of acceptance strength.

Confidence-only produces the larger interface degradation. All-7 accuracy decreases from $0.563$ to $0.482$, mean turns increase from $2.99$ to $5.99$, and ESR falls from $86.3\%$ to $57.8\%$. The corresponding Math-5 turn count rises from $3.42$ to $6.43$. Its lower aggregate PSE does not indicate a superior controller, because confidence-only stops substantially less often and allocates considerably more refinement. Countdown exhibits the complementary failure mode: PSE increases from $5.8\%$ to $17.8\%$ and final accuracy decreases from $0.839$ to $0.659$. Confidence alone can therefore lead either to prolonged refinement or to incorrectly confident termination, depending on the task.

Taken together, the structural and interface ablations show that iterative refinement and the joint verdict--confidence representation provide complementary benefits. The verdict supplies a categorical assessment of the current answer, while confidence determines whether the strength of that assessment exceeds the selected stopping threshold. Requiring their conjunction yields the most consistent aggregate accuracy, compute allocation, and stopping-error behavior across the seven benchmarks.

\subsection{Robustness Across Training Seeds}
\label{app:seed_robustness}

We evaluate the sensitivity of SVR to training randomness by independently training the complete method with seeds 42, 43, and 44. All runs use identical training data, optimization hyperparameters, reward coefficients, prompt templates, and evaluation settings. Adaptive inference uses greedy decoding with $\gamma=0.85$ and $T_{\max}=10$, so the observed variation primarily reflects training stochasticity rather than decoding randomness. Seed 42 is the checkpoint used in the main tables and figures, while seeds 43 and 44 provide additional independent robustness runs. All-7 and Math-5 are computed as unweighted macro-averages over their corresponding datasets. For each metric, we first compute the aggregate value for each seed and then report the mean and sample standard deviation across the three runs.

\begin{table*}[t]
\centering
\small
\setlength{\tabcolsep}{2.2pt}
\renewcommand{\arraystretch}{1.08}
\caption{Aggregate robustness of SVR across three independent training seeds. All runs use adaptive inference with $\gamma=0.85$ and $T_{\max}=10$. Tok. denotes the average cumulative number of prompt and completion tokens consumed per example over all executed turns and is reported in thousands. ESR and PSE are percentages. The final row gives the mean and sample standard deviation across seeds.}
\label{tab:seed_robustness_aggregate}
\begin{tabular*}{\textwidth}{@{\extracolsep{\fill}}lcccccccccc@{}}
\toprule
& \multicolumn{5}{c}{\textbf{All-7}}
& \multicolumn{5}{c}{\textbf{Math-5}} \\
\cmidrule(lr){2-6}
\cmidrule(lr){7-11}
\textbf{Seed}
& \textbf{Acc.}
& \textbf{Turns}
& \textbf{\shortstack{Tok.\\($\times10^3$)}}
& \textbf{\shortstack{ESR\\(\%)}}
& \textbf{\shortstack{PSE\\(\%)}}
& \textbf{Acc.}
& \textbf{Turns}
& \textbf{\shortstack{Tok.\\($\times10^3$)}}
& \textbf{\shortstack{ESR\\(\%)}}
& \textbf{\shortstack{PSE\\(\%)}} \\
\midrule
42
& 0.563 & 2.99 & 8.56 & 86.26 & 29.91
& 0.458 & 3.42 & 11.09 & 82.82 & 36.98 \\
43
& 0.551 & 3.10 & 8.77 & 85.07 & 32.57
& 0.444 & 3.48 & 11.16 & 81.86 & 40.98 \\
44
& 0.555 & 3.06 & 8.70 & 85.14 & 31.97
& 0.445 & 3.57 & 11.39 & 81.46 & 39.94 \\
\midrule
\textbf{Mean $\pm$ Std.}
& $\mathbf{0.556\pm0.007}$
& $\mathbf{3.05\pm0.06}$
& $\mathbf{8.68\pm0.11}$
& $\mathbf{85.49\pm0.66}$
& $\mathbf{31.49\pm1.39}$
& $\mathbf{0.449\pm0.008}$
& $\mathbf{3.49\pm0.07}$
& $\mathbf{11.21\pm0.16}$
& $\mathbf{82.05\pm0.70}$
& $\mathbf{39.30\pm2.08}$ \\
\bottomrule
\end{tabular*}
\end{table*}

Across the independently trained checkpoints in Table~\ref{tab:seed_robustness_aggregate}, both task performance and adaptive compute allocation remain consistent. SVR obtains an All-7 accuracy of $0.556\pm0.007$ and a Math-5 accuracy of $0.449\pm0.008$. The standard deviation of mean inference turns is $0.06$ on All-7 and $0.07$ on Math-5, while cumulative token consumption varies by only $0.11$ thousand tokens on All-7 and $0.16$ thousand tokens on Math-5. All three checkpoints score above the oracle-guided reference in the reported complete-system comparison, so the observed gap is not confined to one favorable training run.

\begin{table*}[t]
\centering
\setlength{\tabcolsep}{2.7pt}
\renewcommand{\arraystretch}{1.04}
\caption{Per-dataset robustness of SVR across three independent training seeds. Final accuracy is reported on the proportional scale. Total Tok. is the average cumulative number of prompt and completion tokens consumed per example over all executed turns and is reported in thousands. ESR and PSE are percentages. Within each metric group, the final row gives the mean and sample standard deviation across seeds.}
\label{tab:seed_robustness_per_dataset}
\begin{tabular*}{\textwidth}{@{\extracolsep{\fill}}llccccccc@{}}
\toprule
\textbf{Metric}
& \textbf{Seed}
& \textbf{Countdown}
& \textbf{GSM8K}
& \textbf{MATH500}
& \textbf{AIME26}
& \textbf{AMC23}
& \textbf{\shortstack{Oly.}}
& \textbf{\shortstack{Minerva}} \\
\midrule
Acc.
& 42
& 0.839 & 0.813 & 0.676 & 0.200 & 0.550 & 0.417 & 0.449 \\
&
43
& 0.828 & 0.807 & 0.674 & 0.167 & 0.525 & 0.412 & 0.441 \\
&
44
& 0.849 & 0.812 & 0.676 & 0.167 & 0.525 & 0.411 & 0.445 \\
&
\textbf{Mean $\pm$ Std.}
& $\mathbf{0.839\pm0.011}$
& $\mathbf{0.811\pm0.003}$
& $\mathbf{0.675\pm0.001}$
& $\mathbf{0.178\pm0.019}$
& $\mathbf{0.533\pm0.014}$
& $\mathbf{0.413\pm0.003}$
& $\mathbf{0.445\pm0.004}$ \\
\addlinespace[2pt]
\midrule
Turns
& 42
& 2.69 & 1.12 & 2.33 & 6.40 & 3.08 & 3.74 & 1.56 \\
&
43
& 2.96 & 1.33 & 2.33 & 6.03 & 3.58 & 3.84 & 1.62 \\
&
44
& 2.55 & 1.03 & 2.27 & 6.20 & 4.12 & 3.77 & 1.49 \\
&
\textbf{Mean $\pm$ Std.}
& $\mathbf{2.734\pm0.212}$
& $\mathbf{1.159\pm0.151}$
& $\mathbf{2.312\pm0.035}$
& $\mathbf{6.211\pm0.184}$
& $\mathbf{3.592\pm0.525}$
& $\mathbf{3.783\pm0.050}$
& $\mathbf{1.555\pm0.066}$ \\
\addlinespace[2pt]
\midrule
\shortstack[l]{Total Tok.\\($\times10^3$)}
& 42
& 3.462 & 1.022 & 4.829 & 23.839 & 7.138 & 16.176 & 3.469 \\
&
43
& 3.924 & 1.701 & 4.837 & 21.975 & 8.630 & 16.725 & 3.619 \\
&
44
& 3.264 & 0.702 & 4.677 & 22.564 & 10.170 & 16.364 & 3.188 \\
&
\textbf{Mean $\pm$ Std.}
& $\mathbf{3.550\pm0.339}$
& $\mathbf{1.142\pm0.511}$
& $\mathbf{4.781\pm0.090}$
& $\mathbf{22.793\pm0.953}$
& $\mathbf{8.646\pm1.516}$
& $\mathbf{16.422\pm0.279}$
& $\mathbf{3.425\pm0.219}$ \\
\addlinespace[2pt]
\midrule
ESR (\%)
& 42
& 89.7 & 100.0 & 90.8 & 60.0 & 85.0 & 80.1 & 98.2 \\
&
43
& 86.2 & 100.0 & 90.8 & 60.0 & 82.5 & 78.2 & 97.8 \\
&
44
& 88.7 & 100.0 & 91.2 & 63.3 & 75.0 & 78.9 & 98.9 \\
&
\textbf{Mean $\pm$ Std.}
& $\mathbf{88.20\pm1.80}$
& $\mathbf{100.00\pm0.00}$
& $\mathbf{90.93\pm0.23}$
& $\mathbf{61.11\pm1.92}$
& $\mathbf{80.83\pm5.20}$
& $\mathbf{79.08\pm0.97}$
& $\mathbf{98.28\pm0.56}$ \\
\addlinespace[2pt]
\midrule
PSE (\%)
& 42
& 5.8 & 18.7 & 23.2 & 40.0 & 30.0 & 38.4 & 53.3 \\
&
43
& 3.5 & 19.6 & 23.4 & 46.7 & 37.5 & 39.9 & 57.4 \\
&
44
& 3.9 & 20.2 & 25.2 & 50.0 & 30.0 & 38.6 & 55.9 \\
&
\textbf{Mean $\pm$ Std.}
& $\mathbf{4.40\pm1.23}$
& $\mathbf{19.48\pm0.72}$
& $\mathbf{23.93\pm1.10}$
& $\mathbf{45.56\pm5.09}$
& $\mathbf{32.50\pm4.33}$
& $\mathbf{38.97\pm0.82}$
& $\mathbf{55.51\pm2.05}$ \\
\bottomrule
\end{tabular*}
\end{table*}

The corresponding dataset-level breakdown in Table~\ref{tab:seed_robustness_per_dataset} further localizes this stability. Final-answer accuracy varies by at most $1.9$ percentage points on every benchmark and by less than $0.5$ points on GSM8K, MATH500, OlympiadBench, and MinervaMath. Countdown also remains consistent despite a modest difference in adaptive compute allocation: seed 44 reaches the highest accuracy of $0.849$ while using the fewest turns and tokens among the three runs. More visible compute variation occurs on GSM8K and AMC23, where a relatively small subset of examples can induce different refinement lengths even when final accuracy remains similar.

Stopping frequency is more stable across seeds than incorrect early stopping. All-7 ESR is $85.49\pm0.66\%$, whereas All-7 PSE is $31.49\pm1.39\%$; the corresponding Math-5 values are $82.05\pm0.70\%$ and $39.30\pm2.08\%$. Thus, independently trained checkpoints learn similar overall stopping frequencies, while the precise subset of incorrectly terminated examples is moderately more sensitive to training randomness. Nevertheless, the small aggregate variations in accuracy, inference turns, token consumption, ESR, and PSE jointly support the reproducibility of the overall accuracy--efficiency behavior of SVR. Because only three independent training runs are available, these results provide robustness evidence rather than a formal statistical-significance analysis.

\end{document}